\documentclass{article}

\usepackage[preprint]{neurips_2025}

\usepackage[utf8]{inputenc} 
\usepackage[T1]{fontenc}    
\usepackage{hyperref}       
\usepackage{url}            
\usepackage{booktabs}       
\usepackage{amsfonts}       
\usepackage{nicefrac}       
\usepackage{microtype}      
\usepackage{graphicx}
\usepackage{subcaption}
\usepackage{amsmath}
\usepackage[table]{xcolor}
\usepackage{geometry}
\usepackage{url}
\hypersetup{
  colorlinks=true,
  linkcolor=blue,        
  citecolor=blue,      
  urlcolor=blue         
}

\definecolor{lavender}{RGB}{230, 230, 250}

\usepackage[most]{tcolorbox}  
\newtcolorbox{methodbox}[1]{
    enhanced,
    title={#1},  
    colback=cyan!5,
    colframe=black,   
    coltitle=white,   
    colbacktitle=yellow, 
    fonttitle=\bfseries,
    attach boxed title to top center={yshift*=0mm}, 
    boxed title style={
        size=small,
        colback=black,
        width=\textwidth, 
        center title,     
    },
    sharp corners,
    width=\textwidth,
    top=0em 
}

\usepackage{colortbl}

\title{Toward Scientific Reasoning in LLMs: Training from Expert Discussions via Reinforcement Learning}

\author{%
  Ming Yin \\
  Princeton University\\
  \texttt{my0049@princeton.edu} \\
  \And
  Yuanhao Qu \\
  Stanford University \\
  \texttt{yhqu@stanford.edu} \\
  \AND
  Ling Yang \\
  Princeton University \\
  \texttt{yangling0818@163.com} \\
  \And
  Le Cong \\
  Stanford University \\
  \texttt{congle@stanford.edu} \\
  \And
  Mengdi Wang \\
  Princeton University \\
  \texttt{mengdiw@princeton.edu} \\
}

\begin{document}

\maketitle

\begin{abstract}
  We investigate how to teach large language models (LLMs) to perform scientific reasoning by leveraging expert discussions as a learning signal. Focusing on the genomics domain, we develop an automated pipeline to extract trainable data and introduce \emph{Genome-Bench}, a new benchmark constructed from over a decade of scientific forum discussions on genome engineering. Our pipeline transforms raw interactions into a reinforcement learning–friendly multiple-choice questions format, supported by 3000+ high-quality question–answer pairs spanning foundational biology, experimental troubleshooting, tool usage, and beyond. We fine-tune an LLM using RL with a rule-based reward signal derived from the synthetic MCQ dataset to enhance domain-specific reasoning. Our results show that reinforcement learning from scientific discussions improves model performance by over 15\% compared to the base model on Genome-Bench, narrowing the gap between open-source LLMs and expert-level reasoning. To our knowledge, this is the first end-to-end pipeline for teaching LLMs to reason from scientific discussions, with promising potential for generalization across scientific domains beyond biology. Our code are available at \url{https://github.com/mingyin0312/RL4GenomeBench}.

\end{abstract}

\section{Introduction}

\begin{figure}[t]
  \centering
   \includegraphics[width=1\linewidth]{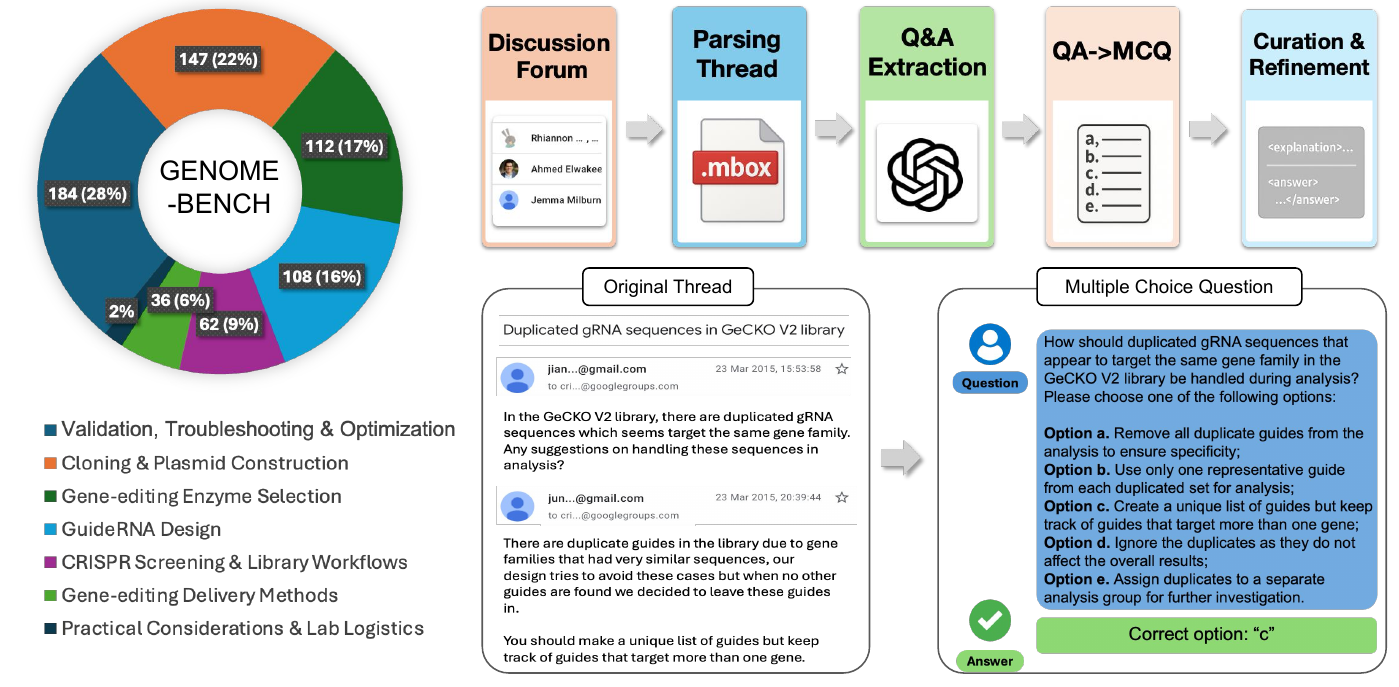}
  \caption{\textit{Overview of Genome-Bench.} We introduce GENOME-BENCH, a novel scientific benchmark comprising 3,332 question-answer pairs derived directly from real-world genome research discussions among human experts over 11 years. The dataset uniquely captures authentic scientific discourse spanning seven key areas—from experimental Validation and Troubleshooting to Practical Considerations and Lab Logistics. Our end-to-end data pipeline begins with parsing raw email threads from archived scientific forums in .mbox format, leveraging a Large Language Model (LLM) to systematically extract question-answer-context tuples. These tuples are then transformed into a structured multiple-choice format, meticulously curated through deduplication, quality filtering, and precise answer annotation to ensure high-quality benchmarking for genomic research applications.
}
  \label{fig1}
\end{figure}

Large language models (LLMs) have shown remarkable capabilities in domains such as mathematics \cite{liu2025understanding,shao2024deepseekmath} and programming \cite{jiang2024survey}. However, despite their impressive performance on standardized reasoning tasks \cite{r1-searcher,chakraborty2024maxmin,guo2024embodied}, LLMs still struggle with expert-level scientific reasoning—particularly in complex, high-stakes fields such as genomics. Unlike mathematical proofs or coding problems, which typically have clear-cut solutions and deterministic evaluation criteria, scientific reasoning often requires interpreting ambiguous observations, accounting for domain-specific nuances, and synthesizing knowledge across multiple layers of abstraction. This gap between current LLM capabilities and the level of reasoning expected from domain experts presents a major challenge for applying LLMs to real-world scientific workflows.

To bridge the gap between LLMs and human expertise, we demonstrate an end-to-end pipeline for training LLMs based on \emph{authentic} scientific discussions among human experts in the field of genomics. As an companion effort detailed in \cite{yin2025genome}, we introduce \textbf{Genome-Bench}, a scientific reasoning benchmark in genomics comprising of questions and answers extracted from over a decade of archived scientific forum discussions among genome editing practitioners \cite{yin2025genome}. These open discussions, drawn from a Google Research forum established in 2013 to explore CRISPR gene-editing technologies, reflect the nuanced, real-world problem-solving and methodological decision-making that working scientists engage in. Unlike curated benchmarks like GPQA \cite{rein2024gpqa}, HLE \cite{phan2025humanity}, and Lab-Bench \cite{laurent2024lab}, which often emphasize standardized testing for LLMs, Genome-Bench captures authentic scientific reasoning as it unfolds in practice. It offers a unique lens into how scientists navigate common misconceptions, interpret ambiguous results, and refine experimental protocols over time. Our findings demonstrate that reinforcement learning and mixture-of-agent training can effectively teach LLMs to emulate this kind of scientific reasoning, bridging the gap between theoretical models and the messy realities of scientific inquiry. Concretely, our main contributions are summarized as follows:

    \begin{itemize}
    
	\item \textbf{Genome-Bench: A novel benchmark grounded in real expert discussions.} We present the first LLM benchmark constructed from real-world scientific discussions, capturing the reasoning processes of practicing researchers. Genome-Bench comprises 3,332 meticulously curated multiple-choice questions derived from over a decade of CRISPR forum archives. Unlike synthetic or exam-based datasets, Genome-Bench reflects the genuine complexity and problem-solving strategies found in frontline experimental biology. The questions span diverse subfields—from experimental troubleshooting and reagent selection to protocol design and tool interpretation—providing a rich and realistic testbed for evaluating domain-specific reasoning in LLMs.
    
	\item \textbf{An end-to-end data pipeline for transforming expert dialogue into RL-compatible data.} We develop an automated end-to-end pipeline that converts raw, noisy discussion email threads into structured question-answer pairs suitable for training and evaluating LLMs with reinforcement learning. This pipeline integrates LLM-assisted question-answer extraction, distractor generation, context preservation, and rigorous quality control to produce high-quality multiple-choice questions. 
    This framework is modular and domain-agnostic, enabling scalable construction of reasoning datasets across scientific domains.
    
	\item 
    \textbf{Enhancing LLM scientific reasoning via reinforcement learning.} We fine-tune the existing LLMs using a rule-based reward signal. Our approach significantly improves their performance on Genome-Bench, achieving over a 15\% accuracy gain compared to instruction baselines. To further elevate the performance, we introduce a reinforcement-learned router model that dynamically selects among multiple fine-tuned expert models for each question. This ``mixture-of-agents'' system yields the best performance and exceeds all the state-of-the-art commercial models. Together, our methods offer a scalable and efficient approach to improving the scientific reasoning abilities of LLMs. 
    
    \end{itemize}

\textbf{Paper Organization.} The remainder of this paper is organized as follows. Section~\ref{sec:related} reviews existing work on related QA benchmarks, domain-specific LLMs, and reinforcement learning for reasoning. Section~\ref{sec:genome} presents Genome-Bench and describes the data extraction pipeline. Section~\ref{sec:rl_router} details our RL fine-tuning methodology and introduces the expert-routing mechanism. Section~\ref{sec:human} compares model outputs to human experts and analyzes the quality of scientific reasoning. The last section concludes the paper.

\section{Related Work}\label{sec:related}

\textbf{Bio-related QA Datasets.} In biomedical and scientific domains, a variety of QA datasets have been curated to facilitate the development and evaluation of domain-specialized models. The BioASQ challenge \cite{krithara2023bioasq} provides a benchmark for biomedical question answering with expert-annotated answers. In the medical domain, PubMedQA \cite{jin2019pubmedqa} targets fact-based questions over biomedical abstracts, MedMCQA \cite{pal2022medmcqa} spans clinical specialties, and MedQA \cite{jin2021disease} and COVID-QA \cite{moller2020covidqa} address disease diagnosis and pandemic-related questions, respectively. These datasets typically derive from formal sources such as exams, medical records, or curated biomedical corpora. More recently, \cite{laurent2024lab} introduced LAB-Bench as a comprehensive benchmark to evaluates LLMs on real-world biology research tasks—such as literature search, protocol planning, and data analysis—offering a more practical alternative to textbook-style QA datasets. In contrast, our work introduces the first QA dataset centered on CRISPR gene editing and constructed from authentic scientific discussions. 

\textbf{LLMs in Scientific Domains.}
Large language models have increasingly been adapted for scientific reasoning by training on domain-specific corpora \cite{zhang2025scientific,prabhakar2025omniscience,liang2024mapping,xie2023darwin,pal2024domain,cai2024sciassess,luo2023biomedgpt,zhang2024generalist,chu20245,wang2025call}. BioBERT \cite{lee2020biobert}, SciBERT \cite{beltagy2019scibert}, and ClinicalBERT \cite{alsentzer2019publicly} are early examples of BERT-based models fine-tuned for biomedical literature and clinical notes. More recent generative models such as BioGPT \cite{luo2022biogpt}, SciFive \cite{phan2021scifive}, and BioBART \cite{yuan2022biobart} extend this trend to sequence-to-sequence tasks. For multimodal tasks, system like GatorTron \cite{yang2022gatortron} combines large-scale LLMs with structured knowledge bases. In general science domain, Galactica \cite{taylor2022galactica} was trained on scientific papers and knowledge graphs to support general scientific writing and inference. In genome engineering, CRISPR-GPT \cite{huang2024crispr} fine-tunes an LLM on gene-editing forum data for experimental design tasks, and BiomedGPT \cite{zhang2024generalist} targets diverse multi-modal biomedical tasks.

\textbf{Reinforcement Learning for LLM Reasoning.}
Reinforcement learning has emerged as a powerful paradigm to enhance models' reasoning capabilities \cite{guo2025deepseekr1,kumar2024training,shinn2024reflexion,guo2025temporal,yang2025reasonflux,wang2024math,r1-searcher,zhong2024dpo,wu2023pairwise,wang2025call,huang2025math}. Reinforcement Learning from Human Feedback \cite{ouyang2022training} utilizes preference data derived from human-annotated comparisons to optimize model behavior via policy optimization methods, notably Proximal Policy Optimization (PPO) \cite{christiano2017deep, schulman2017proximal}. To simplify training, recent work explores rule-based rewards. For example, DeepSeek-R1 devises Group Relative Policy Optimization (GRPO) \cite{shao2024deepseekmath}, which forgoes the PPO value network in favor of group-wise comparisons.
Building upon this, \cite{yu2025dapo} introduced the Dynamic Sampling Policy Optimization (DAPO) algorithm to prevent entropy collapse and improve training efficiency. Similarly, \cite{liu2025understanding} identified optimization biases in GRPO and proposed Dr. GRPO, an unbiased optimization method that improves token efficiency while maintaining reasoning performance. Recently, \cite{xiong2025minimalist} combined rejection sampling and RL for reasoning. Our approach distinguishes itself from prior work by being the first to train LLMs using real scientific forum Q\&A as the primary data source, rather than relying on curated corpora or manually constructed examples.

\section{Genome-Bench}\label{sec:genome}

In this section, we introduce Genome-Bench dataset. An overview is shown in Figure~\ref{fig1}.

\textbf{Data Source and Significance.}
Genome-Bench is a new benchmark dataset designed to evaluate and advance scientific reasoning in large language models, constructed from over a decade of CRISPR-related genome discussions \cite{discussion}. The source material originates from a long-standing public mailing list established by researchers at the Broad Institute of Harvard and MIT. Spanning 2013–2023, this forum captured thousands of real-world scientific inquiries and peer responses related to gene editing. Unlike curated educational datasets or exam-derived corpora, Genome-Bench reflects how scientists actually reason in the lab: asking open-ended, ambiguous questions, proposing competing hypotheses, and resolving experimental uncertainties collaboratively. The resulting dataset comprises of high-quality QA pairs, touching on diverse subfields including foundational biology, experimental troubleshooting, protocol optimization, reagent selection, tool usage, and lab logistics. Because the content is grounded in expert-to-expert conversations, it captures nuanced reasoning patterns, contextual dependencies, and even common misconceptions that synthetic QA datasets often miss.

\textbf{An End-to-End Data Processing Pipeline.}
To transform this corpus into a benchmark suitable for LLM training and evaluation, we developed a fully automated pipeline that processes raw .mbox email archives into structured \emph{multiple-choice questions}. The pipeline includes the following stages:

	1.	\emph{Thread Parsing and Q\&A Extraction.} Each email thread is parsed and preprocessed using a custom parser and GPT-4-Turbo. For each thread, the model extracts a concise scientific question, an expert-provided answer, and supporting context from the surrounding discussion. The result is a structured triplet: (question, answer, context). This ensures that questions retain their real-world framing such as lab-specific constraints and prior experimental attempts.
    
	2.	\emph{MCQ Generation.} The extracted tuples are converted into single-answer MCQs. First, contextual information is prepended to the question to ensure self-containment. Then, the question is rewritten into a fluent, standalone prompt using few-shot GPT-4o instructions. Plausible distractors are generated using LLM prompting, guided to produce scientifically credible but incorrect alternatives that mirror realistic errors. All answer choices are randomly shuffled, with the correct answer explicitly labeled for evaluation or training.
    
	3.	\emph{Answer Encoding and Formatting.} Each item is serialized into a standardized format with consistent labeling. Correct answers are marked (a–e), and expert reasoning is enclosed in <explanation>...</explanation> tags, followed by a final answer tag (<answer>...</answer>). This format facilitates both supervised fine-tuning and reinforcement learning, as it provides both the correct answer and the rationale behind it.
    
	4.	\emph{Quality Filtering.} We applied a multi-stage quality control process. Duplicate or near-duplicate entries were removed, as were questions with ambiguous or trivial answers. Additional filtering excluded off-topic posts and threads lacking substantive expert engagement. The final dataset consists of 3,332 verified, high-quality QA items.

\textbf{Benchmark Structure.} The final 3,332 questions was partitioned into \textbf{a training set} and \textbf{a test set}, comprising 2,671 (80\%) and 661 (20\%) questions, respectively. To enable detailed evaluation of LLM scientific reasoning, each Genome-Bench test question is annotated with two additional fields: \emph{category} and \emph{difficulty}. We defined seven thematic categories based on a corpus-wide analysis: \emph{Validation, Troubleshooting \& Optimization, Cloning \& Plasmid Construction, Gene-editing Enzyme Selection, GuideRNA Design, Screening \& Library Design, Gene-editing Delivery Methods} and \emph{Practical Lab Logistics}. These categories reflect recurring topics in experimental CRISPR work and enable domain-specific performance breakdowns. In parallel, questions are labeled with difficulty levels—\emph{Easy, Medium, or Hard}—based on linguistic structure and conceptual complexity. Heuristics used for this annotation include the presence of conditional clauses, the depth of reasoning required, and the presence of multiple plausible distractors. 
This stratification enables fine-grained evaluation of model robustness as cognitive load increases. We use training data for fine-tuning and test data for evaluation in the following sections.

\section{Enhancing Scientific Reasoning via RL and Expert Routing}\label{sec:rl_router}

\begin{figure}[t]
  \centering
  \includegraphics[width=1\linewidth]{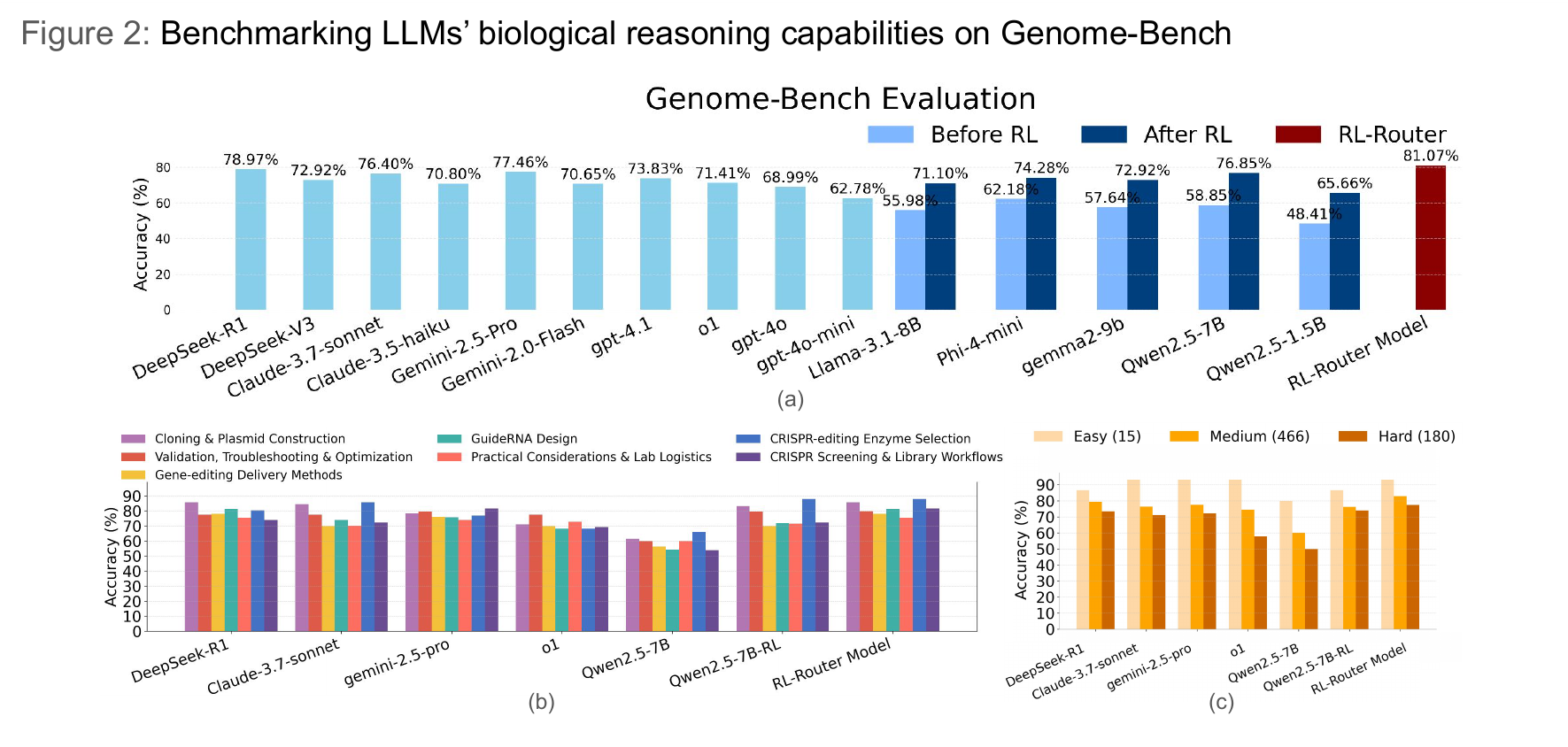}
  \caption{\emph{Benchmarking LLMs’ biological reasoning capabilities on Genome-Bench.} (a) Overall accuracy across Genome-Bench is presented for 21 different LLMs. Smaller models, such as Qwen2.5-7B, show substantial performance improvements when fine-tuned with reinforcement learning (RL), achieving results comparable to leading models like DeepSeek-R1. The RL-Router model further enhances accuracy by dynamically assigning each question to one of four specialized fine-tuned models, selecting the one most likely to provide a correct answer. (b) Comparative accuracy of models across seven biological categories reveals variability in performance. (c) Accuracy is detailed by question difficulty levels—easy, medium, and hard.
}
  \label{fig2}
\end{figure}

\subsection{Reinforcement Fine-Tuning for Scientific Reasoning} 

\textbf{Algorithm and Experimental Setup.} To instill stronger reasoning abilities in LLMs, we fine-tune different open-source models on Genome-Bench using reinforcement learning (RL) and benchmark their performance alongside leading commercial models. The models we evaluate fall into two categories: Base models, which refer to open-source instruction-tuned models, and Commercial models, which represent the current state-of-the-art. Details of these models are summarized in Table~\ref{tab:model-summary}. We refer to the fine-tuned versions of the Base models as RL models.

Specifically, we instantiate RL with Group Relative Policy Optimization (GRPO)~\cite{shao2024deepseekmath}, a sample-efficient alternative to Proximal Policy Optimization (PPO)~\cite{schulman2017proximal} that replaces the traditional value network with group-wise comparisons (check Appendix~\ref{app:grpo}). In GRPO, for each input question $q$, the model generates a group of $G=4$ candidate outputs from the current policy. A reward is assigned to each output based on its correctness and formatting, as detailed below.

\textbf{Reward Function Design.} We employ rule-based  reward design \cite{lambert2024t,liu2025understanding} to align with two objectives of scientific question answering: producing the correct answer and providing extractable format. If the model's response contains a clearly delimited \texttt{<explanation>} followed by a \texttt{<answer>} section, it earns +1 format reward. If the model selects the correct multiple-choice answer (identified by the \texttt{<answer>} tag), it receives +2 correctness reward. Otherwise, it receives 0. A well-structured but incorrect answer receives some credit, and a correct but malformed response avoids full reward. Importantly, the enforced \texttt{<explanation>}+\texttt{<answer>} format encourages model to ground its answer selection based on its reasoning.

\subsection{Experiment Results}
Figure~\ref{fig2} presents a comprehensive evaluation of 21 LLMs on the Genome-Bench test data, revealing key insights into how reinforcement learning (RL) improves scientific reasoning in genomics. In Figure~\ref{fig2}(a), we observe that small open-source models—such as Qwen2.5-7B and Llama3.1-8B—exhibit substantial performance gains (>15\%) after RL fine-tuning, in some cases surpassing commercial baselines like Claude-3.5 and Gemini-2.0. This improvement is also consistently observed in smaller model Qwen2.5-1.5B.
This confirms the effectiveness of reinforcement learning in boosting reasoning capabilities in science domains, especially for models with limited pretraining in scientific contexts. Notably, this finding stands in contrast to prior RL applications in math and coding, which typically involve problems with deterministic answers. 

We further devise the RL-Router model (details in Section~\ref{sec:router}) that achieves the highest accuracy overall (81.07\%), surpassing all individual RL models and even outperforming state-of-the-art commercial systems like DeepSeek-R1. This highlights the advantage of dynamically selecting from a pool of expert models rather than relying on a monolithic architecture.
In Figure~\ref{fig2}(b), accuracy across seven biological categories reveals that the RL-Router maintains strong performance uniformly across domains—ranging from experimental troubleshooting to gene-editing enzyme selection—whereas individual models show domain-specific strengths and weaknesses. Finally, Figure~\ref{fig2}(c) shows that while all models exhibit a decline in accuracy as question difficulty increases, the RL-Router demonstrates strong robustness across easy, medium, and hard tiers. The complete evaluation is deferred to Appendix~\ref{app:category_diff}. Together, these results suggest that RL and expert routing can significantly narrow the gap between open-source LLMs and expert-level scientific reasoning.

\vspace{-0.5em}

\begin{figure}[t]
  \centering
  \includegraphics[width=1\linewidth]{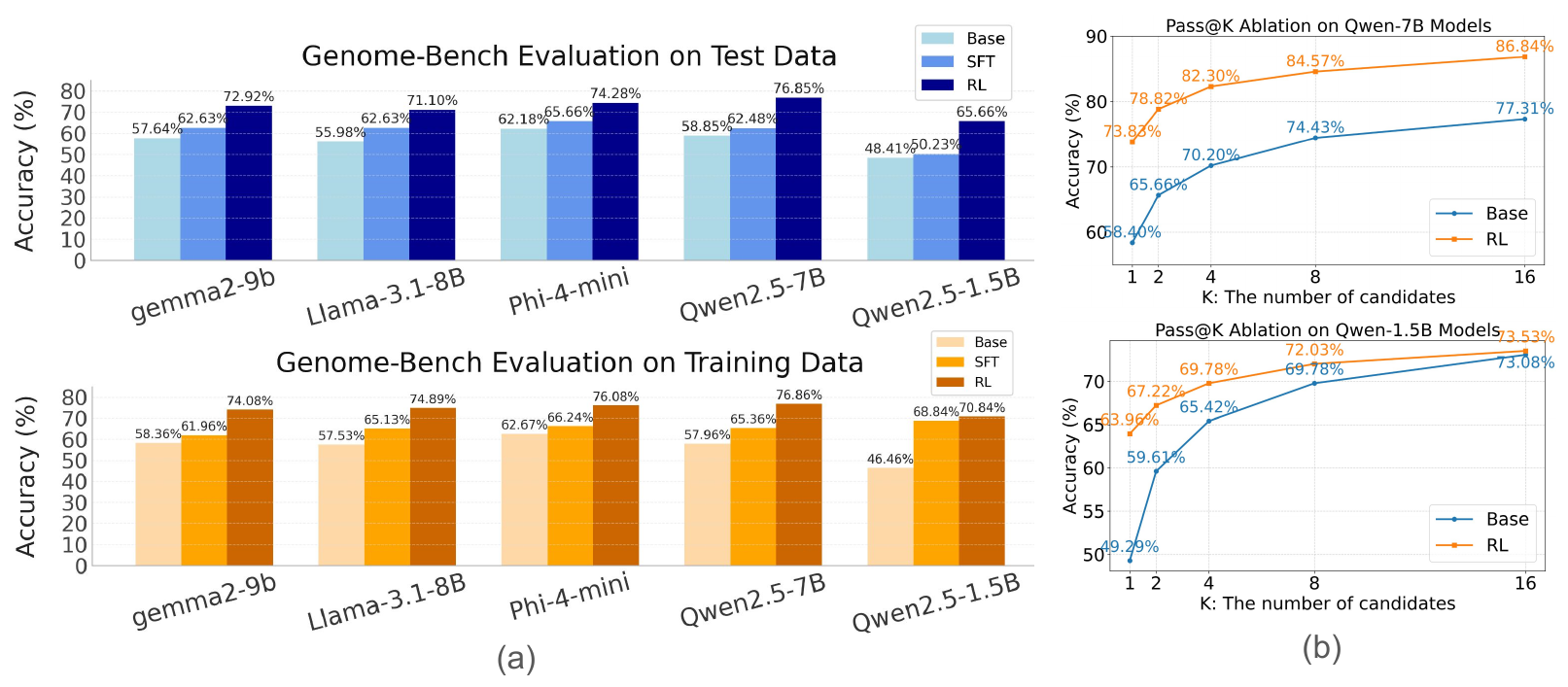}
  \caption{\emph{Ablation Analysis Comparing SFT vs RL.} (a) Comparison of model accuracy across different training stages—Base, SFT (Supervised Fine-Tuning), and RL on both training and test splits. Notably, RL improves generalization on the test set, which is consistent with the performance on the training data. Besides that, RL significantly outperforms SFT. This highlights the advantage of reinforcement learning in aligning model behavior with expert reasoning signals. Moreover, the small model Qwen2.5-1.5B exhibits signs of overfitting during SFT. For instance, the training accuracy under SFT reaches 68.84\%, but the test accuracy drops to just 50.23\%. RL does not exhibit the similar issue. (b) Pass@K evaluation on 7B and 1.5B models. For 7B model, RL method consistently outperforms base models as K increases, suggesting that it acquires new knowledge, as evidenced by the persistent performance gap. In contrast, for the 1.5B model, the performance gap narrows as K increases. This indicates that RL primarily improves the robustness of the output distribution rather than introducing fundamentally new capabilities—likely due to the limited capacity (1.5B).
}
  \label{fig3}
\end{figure}

\textbf{Ablation Analysis: Effectiveness of RL Fine-Tuning.}
Figure~\ref{fig3}(a) shows comparisons across base, SFT, and RL-tuned versions of five LLMs. Reinforcement learning yields significant gains over SFT for all models (>10\%). Besides, RL keeps consistent performance for both train and test, while SFT exhibits overfitting for small (1.5B) model. Figure~\ref{fig3}(b) further demonstrates Pass\@K evaluation, where the 7B and 1.5B RL models reveal different patterns as K increases. For 7B model, RL method consistently outperforms base models as K increases, while for the 1.5B model, the performances between RL and base models cross as K increases. \emph{This finding suggest that the effectiveness of reinforcement learning in helping models acquire and generalize new knowledge might be model-size dependent.} Larger models, such as the 7B variant, appear capable of leveraging RL to refine both accuracy and diversity, while smaller models may lack sufficient capacity to do so. This highlights an important design consideration when applying RL to models of varying scales.

\textbf{Is reinforcement learning enabling accurate reasoning?} 
To assess whether RL improves not only answer accuracy but also the quality of reasoning, we analyzed 10 questions (Appendix~\ref{app:right-wrong}) for which the model initially produced incorrect answers but answered correctly after RL fine-tuning. Our analysis reveals a nuanced picture of RL's impact on reasoning capabilities. On one hand, we observe promising improvements in reasoning paths, such as in question 23, where the RL-tuned model demonstrates correct identification of potential issues raised by the user and provides well-structured logical analysis. However, the results also highlight persistent challenges. For instance, question 7 shows that while RL helps the model arrive at the correct answer, the underlying reasoning remains flawed or incomplete. Similarly, question 27 presents cases where RL successfully guides the model toward the correct answer, but the reasoning remains vague, often simply asserting that a particular choice is best based on context without providing substantive justification. These findings suggest that while RL can effectively improve answer accuracy, ensuring that the model consistently develops sound reasoning processes remains an open challenge and requires further delicate investigation.

\vspace{-0.5em}

\subsection{Router Model with Reinforcement learning}
\label{sec:router}

\begin{figure}[t]
  \centering
  \begin{subfigure}[b]{\textwidth}
    \centering
    \includegraphics[width=\textwidth]{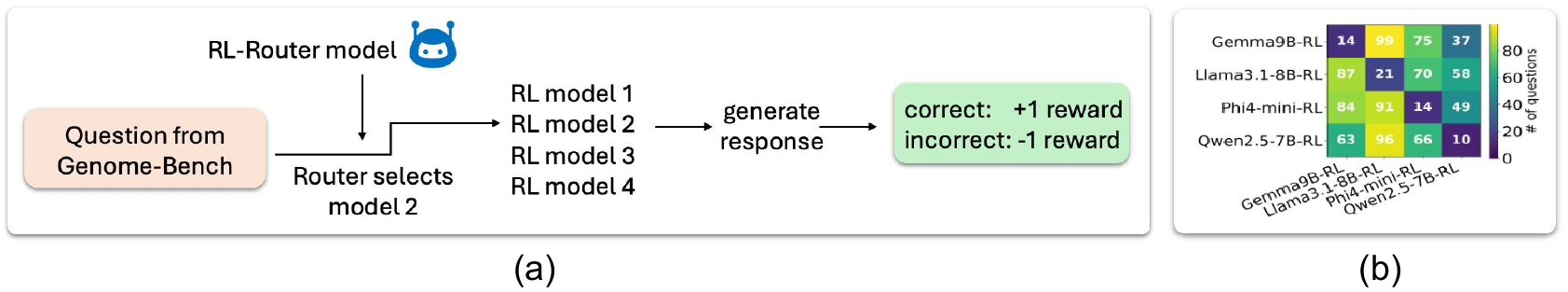}
  \end{subfigure}
  
 
  \begin{subfigure}[b]{\textwidth}
    \centering
    \includegraphics[width=\textwidth]{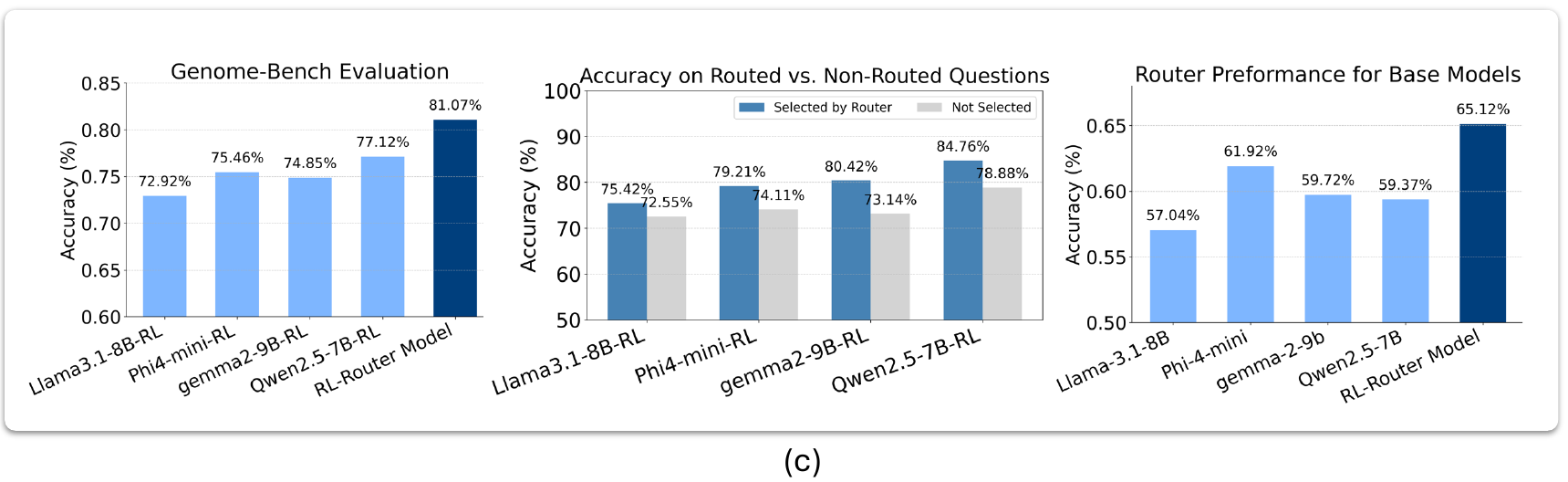}
  \end{subfigure}
\caption{\emph{Reinforcement Learning for Model Routing.} 
(a) Overview of the routing workflow: for each Genome-Bench question, the RL-trained router selects one of four fine-tuned expert models to generate a response, receiving a positive or negative reward based on correctness. 
(b) Pairwise matrix among the RL-tuned experts: diagonal entry $(i,i)$ counts the number of questions only model $i$ answered correctly; off-diagonal entry $(i,j)$ counts questions where model $i$ was correct and model $j$ was not. 
(c) Comparative performance of the RL router. Left: overall Genome-Bench accuracy versus its underlying expert models; Center: accuracy breakdown on routed vs. non-routed questions; Right: ablation results showing router accuracy when selecting among base (non-RL) models, demonstrating gains from routing even without expert fine-tuning. }
 \label{fig4}
\end{figure}

\textbf{RL-Router: Dynamic Model Selection via Reinforcement Learning.} While reinforcement learning enhances individual models, no single model excels at all question types. As shown in Figure~\ref{fig4}(b), the pairwise matrix reveals substantial specialization across the four RL-tuned experts. For instance, among the total 661 test problems, the Llama-3.1-8B-RL model correctly answered 87 questions that Gemma2-9B-RL failed, while Qwen2.5-7B-RL correctly answered 96 questions missed by Llama-3.1-8B-RL. These discrepancies underscore the diversity in model strengths and suggest strong complementarity among the experts—indicating untapped potential if they can be coordinated.

To exploit specialization, we propose a novel ``mixture-of-agents'' system: an \textit{RL-trained router} that dynamically selects the most suitable expert model to answer each question. The router, implemented using Qwen2.5-7B-Instruct, receives the question as input and predicts an index in {1, 2, 3, 4} corresponding to one of the experts as shown in Figure~\ref{fig4}(a). The router is trained via GRPO using a simple reward signal: it receives +1 if the selected expert produces the correct answer, and -1 otherwise. Pre-generated responses from the experts are used to accelerate training. Over time, the router learns to recognize patterns in question content and domain that correlate with each expert's strengths.

\textbf{Experiment Results.}
Figure~\ref{fig4}(c) shows that the RL-Router consistently outperforms all individual RL-tuned models, achieving a new state-of-the-art accuracy of 81.1\% on the Genome-Bench test set. Such an improvement arises from the router’s ability to dynamically allocate questions to the expert model best suited for each query—as further verified in the center panel of Figure~\ref{fig4}(c), where routed questions consistently yield higher accuracy than non-routed ones. The routing distribution is provided in Figure~\ref{fig:rl_router_7B_rl}. Together with Figure~\ref{fig2}(a), the RL-Router also exceeds the performance of top-tier commercial systems such as DeepSeek-R1. Beyond aggregate accuracy, Figures~\ref{fig2}(b) and \ref{fig2}(c) demonstrate that routing leads to robust improvements across all the scientific categories and difficulty levels. This adaptability underscores the router’s ability to make fine-grained, context-aware decisions comparing to the existing models.

\textbf{Ablation Analysis: Understanding Routing Efficacy.}
We conduct ablations to isolate the router's impact. First, we evaluate a router trained over base (non-RL) models (Figure~\ref{fig4}(c),left;Figure~\ref{fig:rl_router_7B_base}). Even without RL-enhanced experts, the router improves overall accuracy, highlighting the inherent benefit of expert selection. Second, we experiment with a smaller router (Qwen2.5-1.5B, Figure~\ref{fig:rl_router_1.5B}). This underpowered router converges to favoring a single expert, failing to learn nuanced routing policies. As a result, accuracy gains are limited. This phenomenon suggests that routing itself is a learning task requiring sufficient model capacity and that deploying a capable LLM as the router is crucial for harnessing the full benefits of specialization. More details are deferred to Appendix~\ref{app_sec:router}.

In summary, our combination of reinforcement fine-tuning and RL routing yields state-of-the-art performance on Genome-Bench. This framework offers a scalable path toward scientific LLMs that both reason accurately and know when to defer to the most competent expert.

\section{Human Evaluation Analysis}\label{sec:human}

\begin{figure}[t]
  \centering
  \begin{subfigure}[b]{0.45\textwidth}
    \centering
    \includegraphics[width=0.8\textwidth]{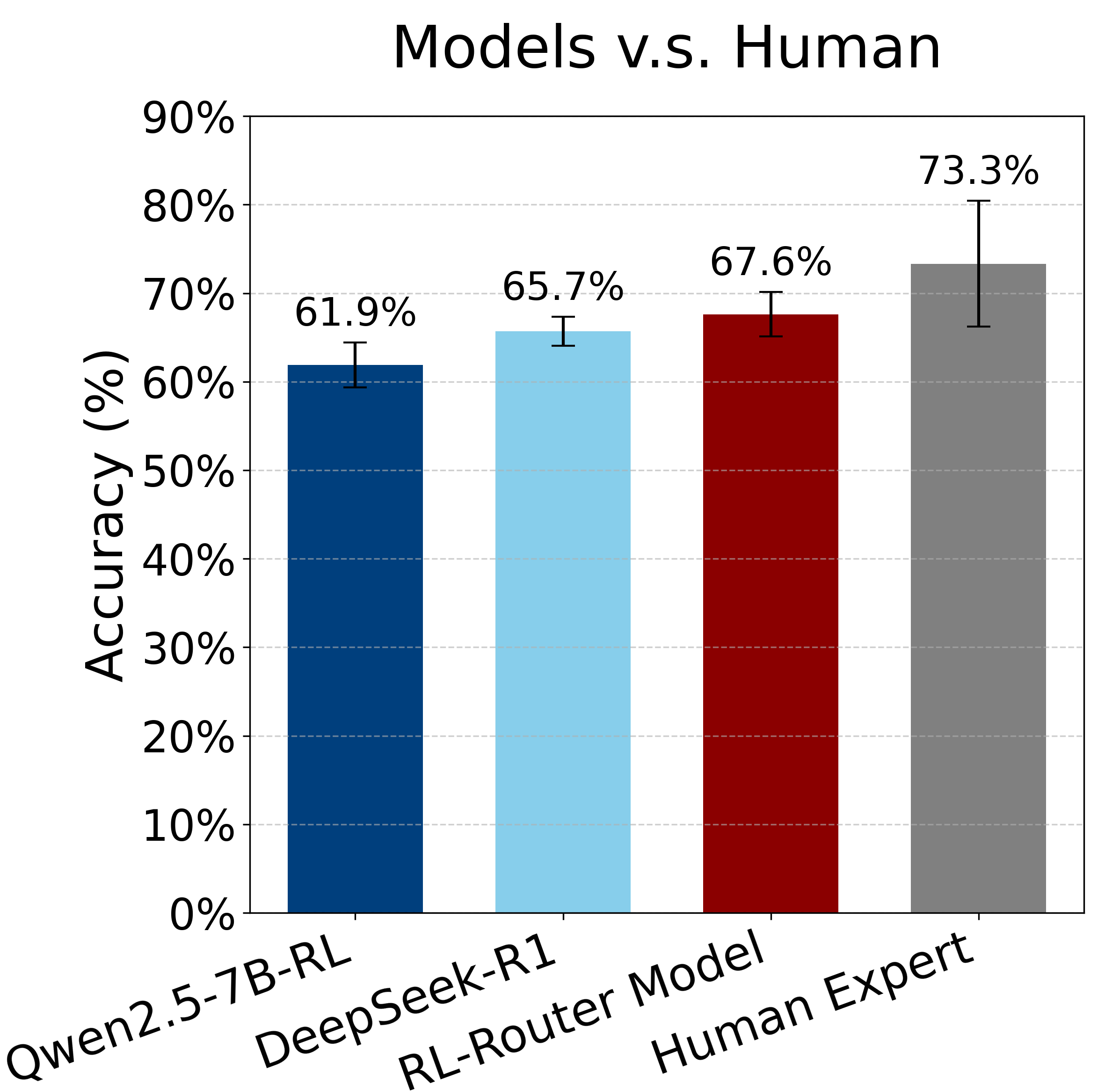}
    \caption{}
    \label{fig5:1}
  \end{subfigure}
  \hfill
  \begin{subfigure}[b]{0.45\textwidth}
    \centering
    \includegraphics[width=0.9\textwidth]{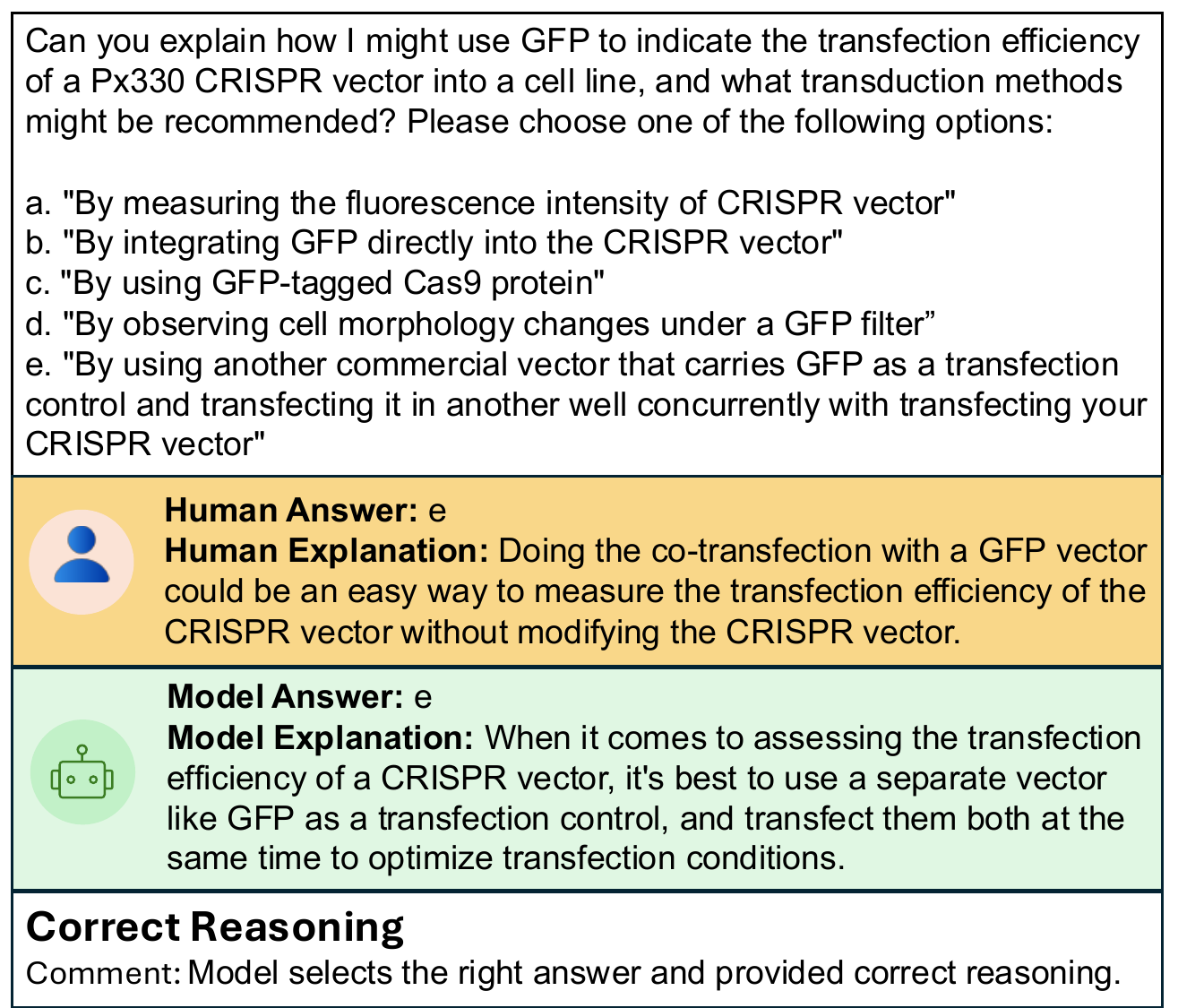}
    \caption{}
    \label{fig5:2}
  \end{subfigure}

  \begin{subfigure}[b]{0.45\textwidth}
    \centering
    \includegraphics[width=0.9\textwidth]{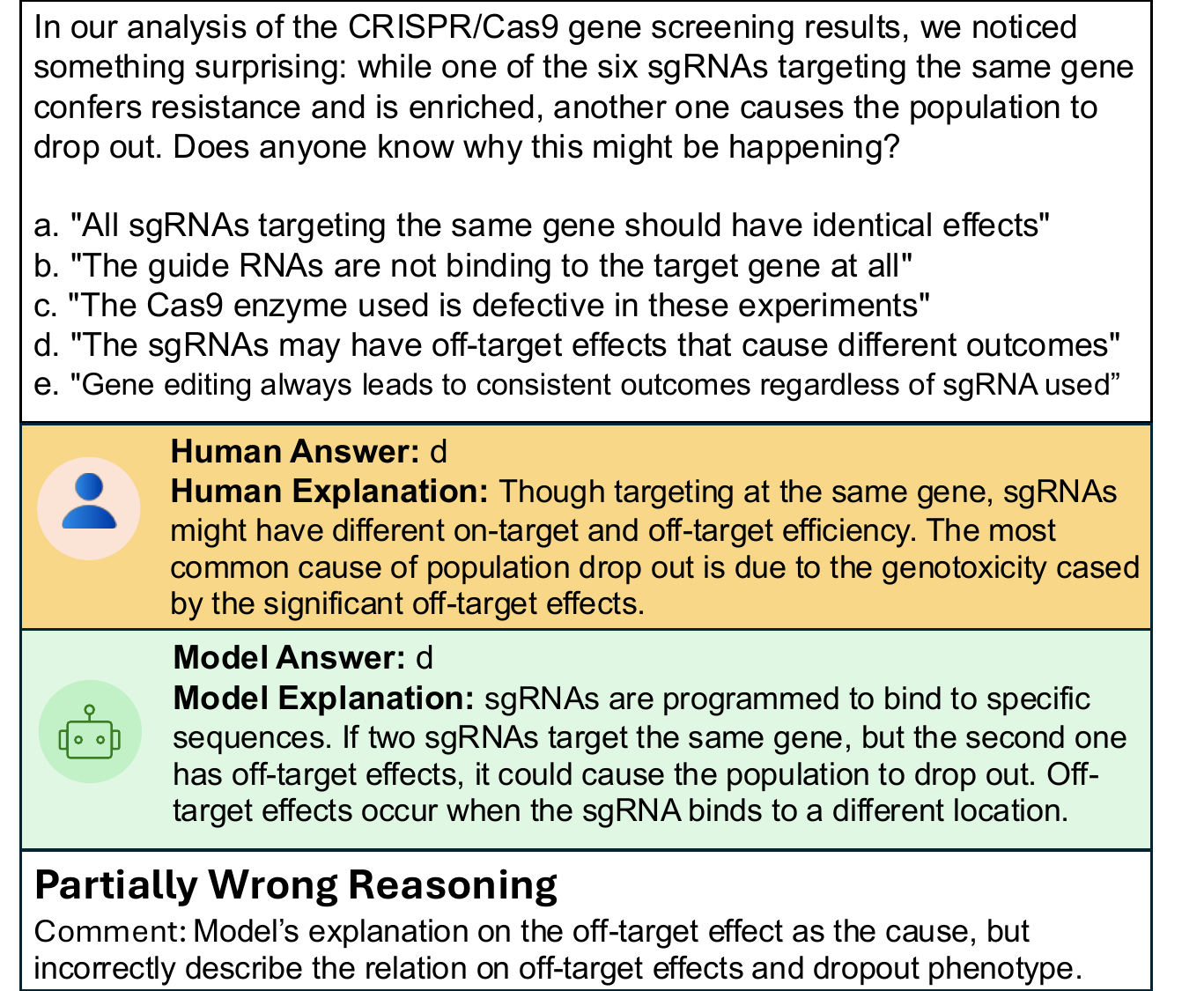}
    \caption{}
    \label{fig5:3}
  \end{subfigure}
  \hfill
  \begin{subfigure}[b]{0.45\textwidth}
    \centering
    \includegraphics[width=0.9\textwidth]{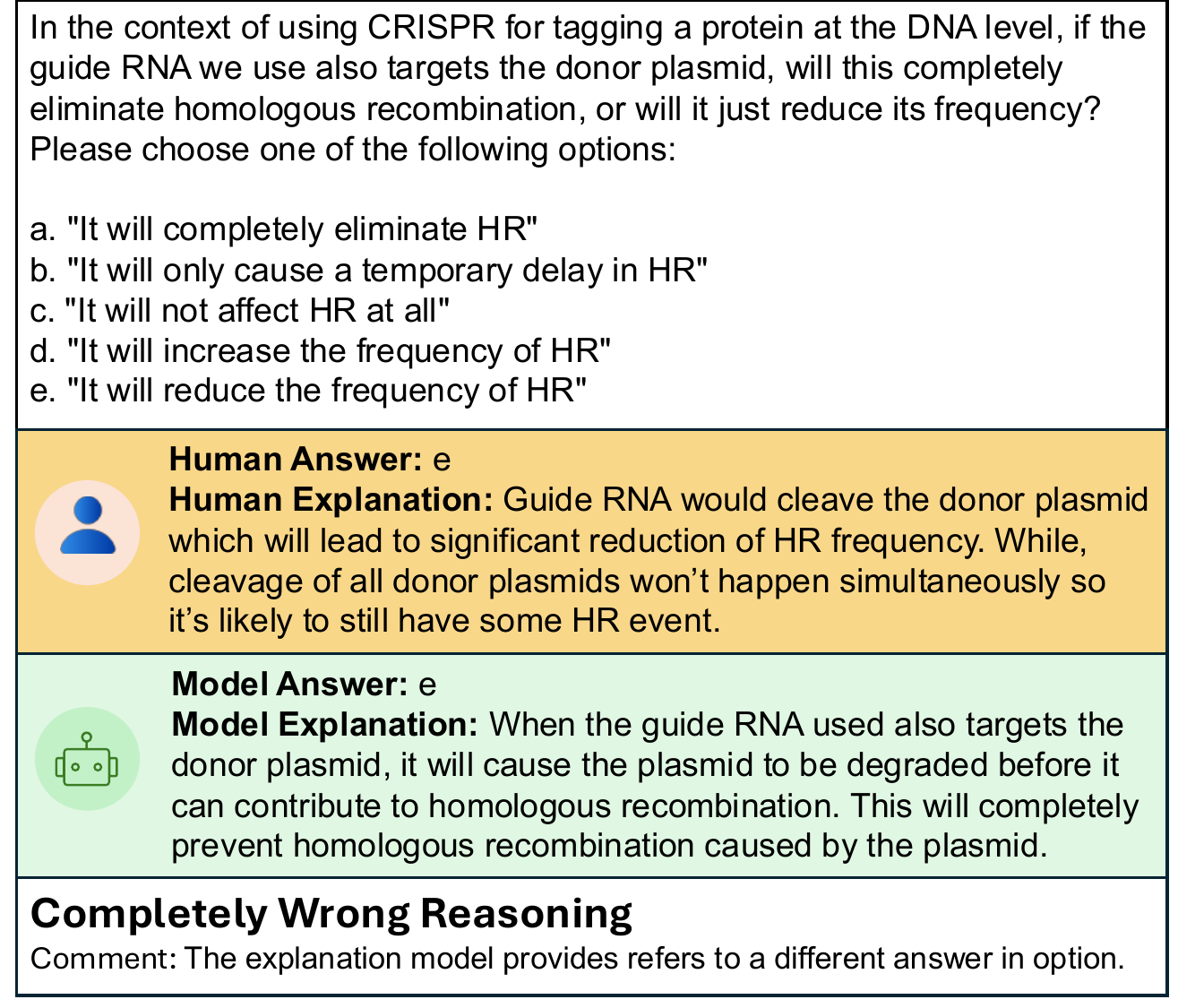}
    \caption{}
    \label{fig5:4}
  \end{subfigure}
\caption{\emph{Human Evaluation and Analysis of Model Reasoning.} Accuracy comparison on 35 expert-curated questions, evaluated across Qwen2.5-7B-RL, DeepSeek-R1, the RL-Router model, and human experts. The RL-Router model achieves the highest model accuracy (67.6\%), approaching the performance of human experts (73.3\%). Qualitative analysis of RL-Router model-generated explanations compared to human-provided reasoning. Each example is categorized as \emph{Correct Reasoning}, \emph{Partially Wrong Reasoning}, or \emph{Completely Wrong Reasoning} based on the scientific soundness and alignment with the ground-truth rationale.}
\label{fig5}
\end{figure}
    
\textbf{Human Evaluation Design.}
To rigorously assess the LLM's scientific reasoning ability, we conducted a controlled human expert evaluation using a curated subset of questions from Genome-Bench. We selected 35 challenging and diverse items spanning a range of biological topics with difficulty levels being Medium or Hard. These questions were chosen to reflect realistic problems encountered in genome engineering practice. We recruited three domain experts in CRISPR and molecular biology. Each expert was instructed to complete the 35 questions in 60 minutes without the use of external sources. The format contains a justification enclosed in <explanation> tags, followed by a selected option in <answer> tags. This design not only facilitated direct comparison between human and models, but also preserved interpretable reasoning traces. The resulting human annotations serve as a human expert reference set, enabling both quantitative benchmarking and qualitative assessments.

We further posed the same 35-question set to three different models:
	1. Qwen2.5-7B-RL, our strongest single model fine-tuned using RL;
	2. DeepSeek-R1, a top-performing baseline commercial LLM;
	3. The RL-Router model, our mixture-of-agents system.
Each model was queried in three independent runs, and all model outputs adhered to the <explanation>/<answer> structure. 

\textbf{Quantitative Results.}
In Figure~\ref{fig5}(a), the average human expert accuracy across the 35 questions was 73.3\%, establishing a baseline for expert-level reasoning. Among the models, the RL-Router achieved 67.6\% accuracy, the highest among all tested LLMs and the closest to human expert performance with the gap 5.7\%. As a comparison, DeepSeek-R1 scored 65.7\%, and the Qwen2.5-7B-RL achieved 61.9\%. These results demonstrate that our reinforcement-tuned routing approach significantly narrows the performance gap between open-source LLMs and human experts, highlighting the effectiveness of RL training using structured discussion data in combination with routing strategies.

\textbf{Qualitative Analysis.}
Beyond final answer accuracy, we conducted a fine-grained qualitative evaluation of the model's reasoning chains in Figure~\ref{fig5}(b)-(d). Each model-generated explanation was evaluated by the human expert and categorized into one of three reasoning types.

\emph{Correct Reasoning \ref{fig5:2}:} The model selected the correct answer and provided a scientifically coherent explanation that mirrored expert reasoning. For instance, in a question regarding the use of GFP as a transfection control, the model and human both identified co-transfection with a GFP vector as the correct strategy and provided justifications rooted in standard laboratory practice. These cases illustrate that the model has internalized domain-appropriate problem-solving logic.

\emph{Partially Wrong Reasoning \ref{fig5:3}:} The model reached the correct answer, but its explanation contained factual inaccuracies, omissions, or speculative logic. For example, in a question about sgRNA screening outcomes, the model selected the right answer (“off-target effects”) but mischaracterized the biological mechanism. These instances suggest that the model may arrive at the correct solution by chance or by leveraging surface-level cues, rather than engaging in deep causal reasoning.

\emph{Completely Wrong Reasoning \ref{fig5:4}:} The model’s explanation failed to reflect valid scientific reasoning, even when the final answer happened to be correct. As an example, when asked about the impact of guide RNA self-targeting on homologous recombination, the model claimed the process would be completely blocked—contradicting to its own selection of the answer ``e'', which correctly noted only a reduction in recombination frequency. These failures expose the fragility of LLM reasoning fidelity and the limitations of current training in capturing causal understanding.

\textbf{Discussion and Implications.} This evaluation yields two core findings. First, reinforcement-trained LLMs are rapidly approaching expert-level accuracy on complex, domain-specific scientific questions. The RL-Router model in particular demonstrates that careful integration of domain-specific data, structured reasoning supervision, and dynamic expert selection can bring open-source LLMs within striking distance of human expert performance. Second, and equally important, reasoning fidelity remains an open challenge. While models increasingly get the right answers, their reasoning pathways may still diverge significantly from expert logic. This presents risks for downstream scientific applications, where valid justifications are as critical as correct predictions. Our findings reinforce the need to supervise the reasoning process explicitly, not just the answer outcome. Collectively, this evaluation highlights the significant progress made by RL-trained LLMs in scientific reasoning. Figure~\ref{fig5} serves both as a testament to current capabilities and as a diagnostic lens, revealing areas where models still fall short in the step-by-step reasoning that characterizes human scientific thought.

\section{Conclusion and Limitations}

We present an end-to-end framework for LLMs to perform scientific reasoning by training on real-world expert discussions. We introduce Genome-Bench, a benchmark comprising over 3,000 multiple-choice questions from CRISPR-related forums, and apply reinforcement fine-tuning with a structured reward signal to achieve over 15\% accuracy gains compared to existing models. Our RL-Router model, which dynamically selects among specialized experts, further attained 81.1\% accuracy on Genome-Bench. While RL improves answer accuracy, qualitative analysis reveals that reasoning remains imperfect, with some explanations lacking causal or scientific rigor. Future work may explore hybrid reward mechanisms that penalize inconsistent reasoning, leverage human-in-the-loop critiques, or align LLM explanations with mechanistic scientific principles. We leave these as future works.





\bibliographystyle{plain}
\bibliography{ref}

\clearpage

\clearpage
\appendix

\section*{ \centering Appendix}

\section{ Detailed pipeline for Constructing the Genome-Bench Dataset}

\begin{figure}[htbp]
  \centering
  \begin{subfigure}[b]{0.41\textwidth}
    \centering
    \includegraphics[width=\textwidth]{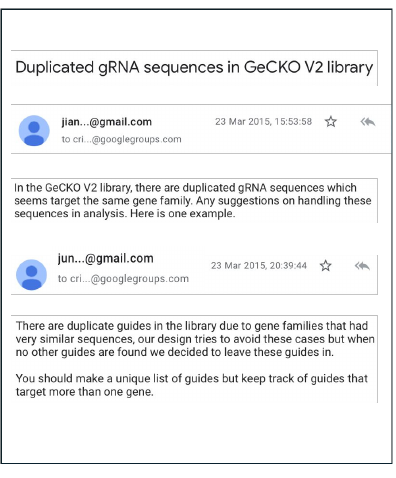}
    \caption{Step 1: Original Google Group Thread}
    \label{fig:appendix1}
  \end{subfigure}
  \hfill
  \begin{subfigure}[b]{0.45\textwidth}
    \centering
    \includegraphics[width=\textwidth]{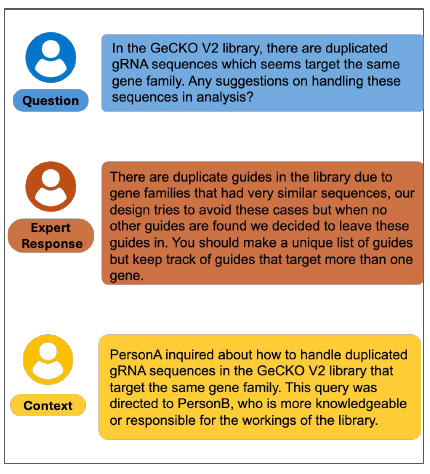}
    \caption{Step 2: Extract (Question, Answer, Context)}
    \label{fig:appendix2}
  \end{subfigure}

  \vspace{0.5em}

  \begin{subfigure}[b]{0.42\textwidth}
    \centering
    \includegraphics[width=\textwidth]{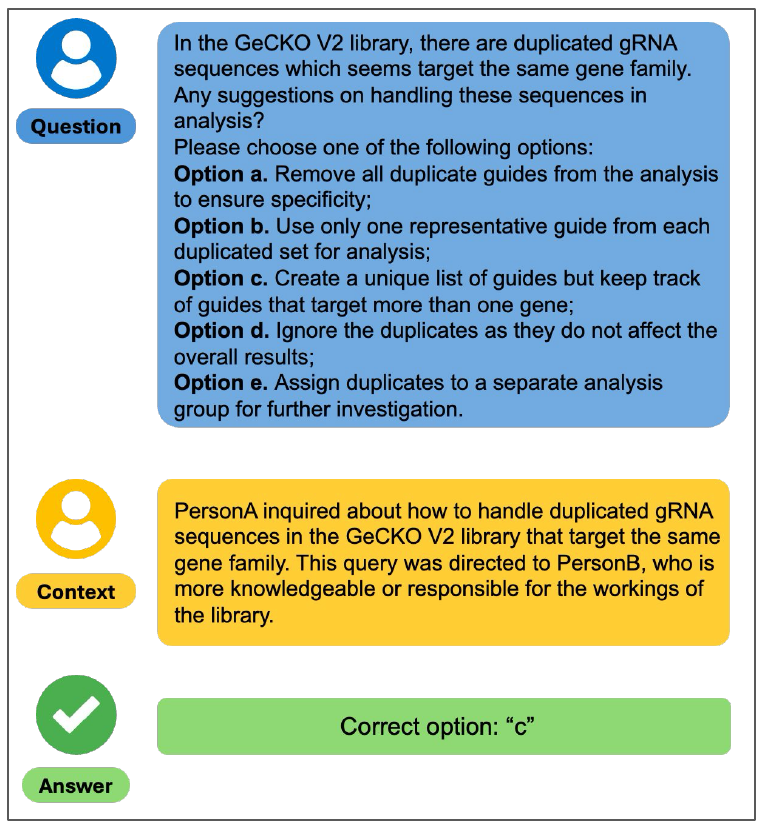}
    \caption{Step 3: Generate Multiple Choices}
    \label{fig:appendix3}
  \end{subfigure}
  \hfill
  \begin{subfigure}[b]{0.44\textwidth}
    \centering
    \includegraphics[width=\textwidth]{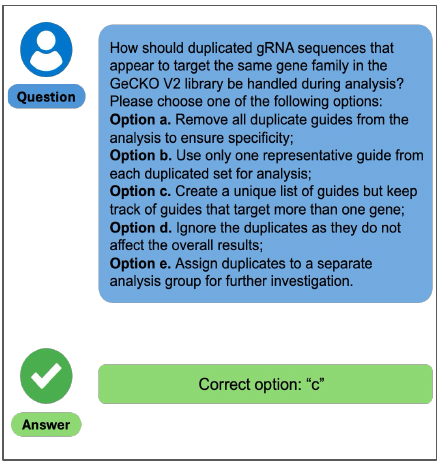}
    \caption{Step 4: Finalize Question-Answer pairs}
    \label{fig:Finalize Question-Answer pairs}
  \end{subfigure}

  \caption{A Step-by-step Example of Genome-Bench Creation. This figure illustrates the data extraction pipeline for curating the Genome-Bench. Initially, the original discussion thread is parsed into a structured format consisting of a Question, Expert Response, and Context, where the context is carefully crafted to reduce potential hallucinations. The expert's response is subsequently transformed into a multiple-choice question, with the correct answer directly derived from the expert's original reply. Finally, the QA pair is refined by integrating the context into the initial question.
}
  \label{fig:crisprbench_pipeline}
\end{figure}

\section{Genome-Bench Creation Pipeline}

\subsection*{Data Source} 
The source of our dataset is an open, public discussion forum ``Genome Engineering using CRISPR/Cas Systems,'' \cite{discussion} initially established by the Feng Zhang lab at the Broad Institute of MIT and Harvard. This forum served as a dynamic, crowd-sourced Q\&A platform where scientists worldwide could post questions about CRISPR gene-editing tools and laboratory practices. Over 11 years, it amassed a wealth of inquiries and expert responses, culminating in approximately 4,000 discussion threads.

\subsection*{Data Parsing and Q\&A Extraction}
The raw dataset, exported in \texttt{.mbox}, is parsed and converted into DataFrame format using Pandas, where each row corresponds to an email thread. Each unique email thread is individually pre-processed by OpenAI’s GPT-4 Turbo model and reformatted for the purpose of fine-tuning. The model is tasked with extracting Q\&A pairs by interpreting the textual content of each thread. The model is prompted to process the current email thread and identify scientific and research related questions and answers (Q\&A Pairs). Because certain Q\&As are specific and context-driven, the model is prompted to use the entire thread to provide a ``context'' field for each Q\&A Pair. It finally outputs structured data with the following format: \texttt{\{question, answer, context\}}. The Q\&A dataset is anonymized after processing.

\subsection*{Q\&A Format to Multiple-Choice Question (MCQ) Transformation}
To utilize the data for training and evaluation, we converted each Q\&A pair into a structured multiple-choice question format. The transformation was carefully designed to preserve the original problem and answer while adding plausible alternative options to form a challenging MCQ. Our pipeline consists of several sequential steps:

\textbf{Context Integration:} Many forum questions rely on context from previous messages (e.g., experimental conditions, what has been tried already). We prepend any available context to the question to make it self-contained. For example, if a question was asked in a thread after a description of an experiment, we combine them: \texttt{``Question context: ... [context description] ... Question: ... [actual question]''}. This ensures that important background information is not lost.

\textbf{Question Rewriting:} We then convert the question (with context) into a natural, stand-alone question that reads clearly on its own. This involves rewriting to integrate the context smoothly and remove any forum-specific phrasing (such as direct references to users like ``Person A''). We employed a strong LLM (GPT-4o in our case) to perform this rewriting via prompt-based instruction. The prompt asked the model to preserve all factual information while phrasing the question in a single, coherent sentence or paragraph. For instance, the structured input \texttt{"Question context: ... PersonB suggests ... PersonA addresses. Question: Is it possible your Amp/Carbenicillin has gone bad?"} was rewritten by GPT-4 to: \texttt{“Could the issue with your lentiCrispr v2 growth after cloning the gRNA be due to your ampicillin or carbenicillin going bad?”} – a fluent question that incorporates the context. This step yields a polished question ready for future steps.

\textbf{Distractor Generation:} Next, we generate four distractor options (incorrect answers) to accompany the correct answer. The correct answer is derived from the original answer text or its key conclusion. We prompt GPT-4o with the question and the correct answer (explanation) to produce plausible alternative answers that could be chosen by a non-expert or a confused model. The model is instructed to produce options that are credible yet definitively incorrect given the context. For example, if the correct answer explains that a plasmid’s large size likely caused a transfection failure, distractors might include other causes like reagent issues or claims that plasmid size doesn’t matter. The generation is done in a prompt that asks for five candidate answers (one being the true answer, and four false ones). We then identify which of the model’s outputs corresponds to the known correct answer and label it accordingly. This step yields a set of five answer statements (a–e).

\begin{figure}[t]
  \centering
  \includegraphics[width=0.8\linewidth]{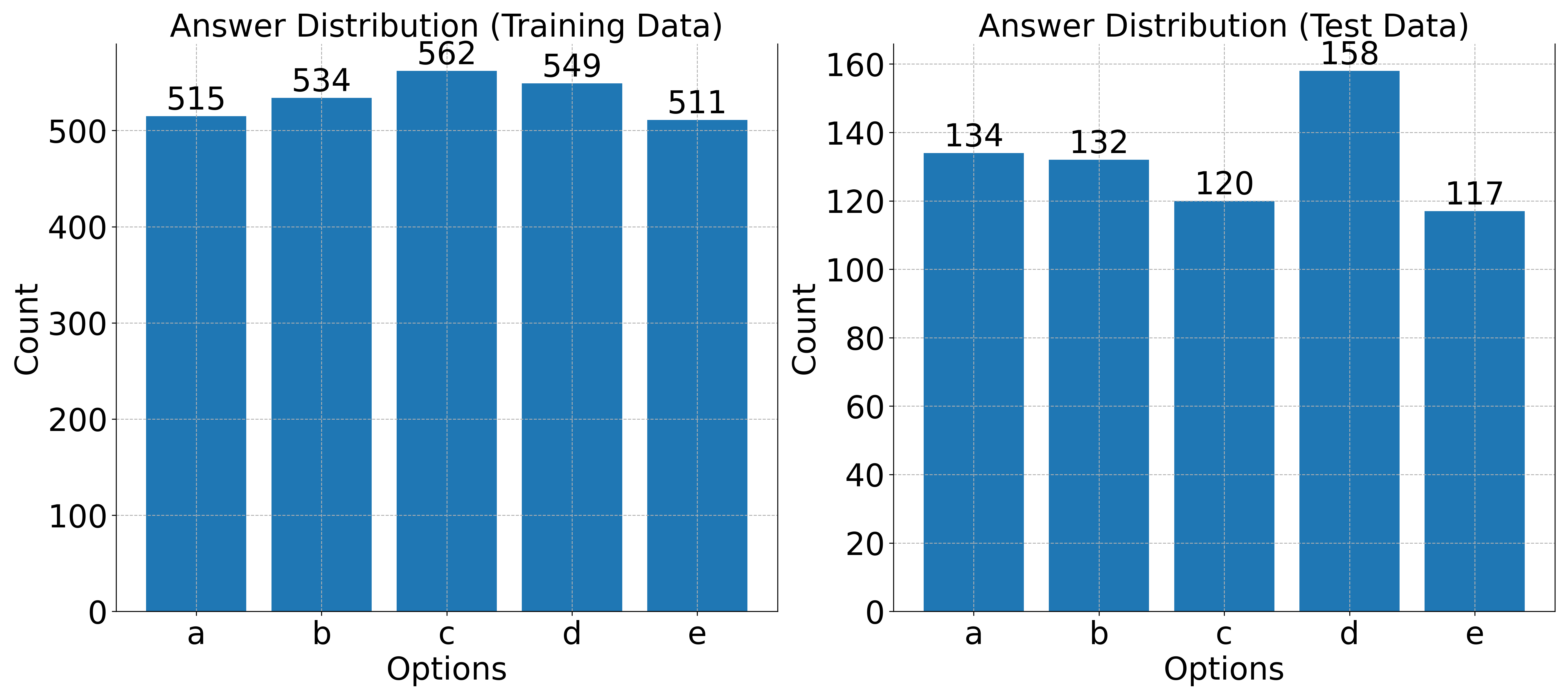}
  \caption{Answer distribution for Genome-Bench.}
  \label{fig:an_di}
\end{figure}

\textbf{Option Formatting and Shuffling:} We format the answer options as lettered choices prefixed by \texttt{a., b., c., d., e.}, and ensure they follow a consistent style, and we randomize the order of the options. The correct answer’s label is tracked during shuffling so that we know which letter is correct after permutation. Shuffling prevents positional bias and makes the dataset more robust (see Figure~\ref{fig:an_di}). We merge the rewritten question and the shuffled options into one consolidated question text. In the final dataset, the question field contains both the prompt and the available choices. For example, \texttt{“Question: Could the issue with your lentiCrispr v2 growth… be due to your Ampicillin or Carbenicillin going bad? Please choose one of the following options: a. … b. … c. … d. … e. …”}. This step ensures each data entry is a complete multiple-choice question ready to be posed to a model or human.

\textbf{Answer and Explanation Encoding:} We construct an answer field containing the correct option label and the explanation. We wrap the original answer explanation (usually a few sentences of reasoning from the expert) in \texttt{<explanation>...</explanation>} tags, and we place the correct option letter in \texttt{<answer>...</answer>} tags. For instance, an entry’s answer might be: \texttt{<explanation>Yes, larger plasmid size can reduce transfection efficiency in certain cell lines, so using too much DNA likely caused the issue.</explanation> <answer>e</answer>}. The explanation text is taken directly from the expert’s answer, ensuring we preserve the reasoning behind the correct choice.

After these steps, each data sample is a JSON object with a question (including integrated context and all answer options) and an answer (containing the correct choice and explanation). The dataset is thus in a structured QA format suitable for both training an LLM (where the model is given the question and must output the answer) and evaluating its multiple-choice accuracy. An example from our training set is shown below:

\begin{quote}
\textbf{Question:} In the process of using CRISPR technology on plants, particularly when considering whether these plants can be classified as cisgenic, isn't there a possibility that some T-DNA sequence from the plant binary vector backbone might insert into the plant genome during Agrobacterium-mediated transformation?\\
Please choose one of the following options:
\begin{itemize}
  \item a. Only in rare cases, the T-DNA sequence from the binary vector may insert into the plant genome.
  \item b. T-DNA sequences are never inserted into the plant genome during transformation.
  \item c. The plant genome naturally resists any insertion of T-DNA sequences from the binary vector backbone.
  \item d. Yes, the T-DNA sequence from the binary vector backbone typically inserts into the plant genome.
  \item e. No, T-DNA insertion is highly selective and does not include the binary vector backbone.
\end{itemize}

\textbf{Answer:} {<explanation>}Yes, during Agrobacterium-mediated transformation, some T-DNA from the plant binary vector backbone typically inserts into the plant genome.{</explanation>} {<answer>d</answer>}
\end{quote}

\subsection*{Quality Control}
Because the raw mined MCQ can be noisy, we performed several quality control and filtering steps to clean the dataset before use:

\textbf{Deduplication:} We removed duplicate or nearly identical questions to avoid redundancy. Some questions were asked multiple times over the years; these were identified (using string matching and manual verification) and only one representative instance was kept.

\textbf{Incomplete/Unanswered Removal:} We filtered out any questions that did not have a clear answer in the discussion. Specifically, entries where the extracted answer was empty, merely a confirmation, or contained phrases indicating no answer (e.g., “No answer”, “unanswered”, “no response”) were discarded. In total, we removed on the order of 180+ Q\&A pairs in this category, which had no informative value due to lack of resolution.

\textbf{Low-Quality Filtering:} We also pruned questions that were overly vague, off-topic, or lacking scientific substance, as identified by human inspection. These included generic pleas for help (“Can someone please give suggestions what could be going wrong?”), questions about website availability and collaboration requests.

After cleaning, we arrived at a finalized set of 3,332 single-choice multiple-choice questions, which we split into a training set and a test set. We reserved 20\% for testing, resulting in 2,671 training questions and 661 test questions. An additional example is also visualized in Figure~\ref{fig:crisprbench_pipeline}.

\subsection*{Genome-Bench Evaluation Protocol}
To ensure consistency across all evaluations, we employed a structured system prompt that specifies the required response format. The model is instructed to first generate an internal reasoning process enclosed within \texttt{<explanation>...</explanation>} tags, followed by its final answer enclosed within \texttt{<answer>...</answer>} tags. The final answer must be exactly one of five choices: \texttt{\{a, b, c, d, e\}}. This format ensures that models not only select an answer but also provide traceable reasoning steps.

\textbf{Inference Setup:} We utilized the vLLM library to efficiently run inference with large language models. Specifically, each question was provided to the model along with the system prompt and task-specific instructions. Sampling parameters (e.g., temperature, top-p) were kept default for all evaluations for fair comparisons. The outputs were collected for all 661 test questions.

\textbf{Answer Extraction:} After inference, each generated output was parsed to extract the final answer. We implemented a simple parser that locates the text enclosed within the <answer>...</answer> tags. In cases where the expected format was not adhered to, the output was considered invalid and treated accordingly during evaluation.

\section{Methodological Details of Reinforcement Fine-tuning}\label{app:grpo}

In the reinforcement fine-tuning stage, we leverage the \emph{Group Relative Policy Optimization} (GRPO) algorithm \cite{shao2024deepseekmath} to further improve the model. GRPO is a variant of \emph{Proximal Policy Optimization} (PPO) \cite{schulman2017proximal} tailored for large language model training. The key idea of GRPO is to forego a separate value (critic) network and instead estimate advantages by comparing a group of sampled responses for the same prompt. This design is motivated by the limitations of standard PPO in our setting: PPO requires learning a value function to supply baselines for advantage estimation, which doubles the model’s memory footprint. By contrast, GRPO computes a baseline reward on-the-fly from a group of candidate outputs, significantly reducing the extra parameters and training overhead. 

\subsection*{GRPO Objective Function}

Formally, let $q$ denote an input query (prompt) sampled from the dataset. For each $q$, we sample a group of $G$ candidate outputs $o_1, o_2, \ldots, o_G$ from the current policy (treated as the ``old'' policy for the update). The reward model then assigns each output $o_i$ a scalar score $r_i$.

Instead of using a learned value function, we compute a \textbf{baseline} reward from this group by taking the average score:

\[
\bar{r} = \frac{1}{G} \sum_{i=1}^{G} r_i.
\]

The \textit{advantage} of each output $o_i$ is defined as its reward minus the baseline. In practice, we normalize this advantage by the group's standard deviation $\sigma_r$ for stability. Thus the advantage for output $o_i$ (applied to all tokens in that output) is:

\[
A_i = \frac{r_i - \bar{r}}{\sigma_r},
\]

as used in the outcome-based setting. Intuitively, each sampled response’s reward is judged \textit{relative} to its peers: outputs scoring higher than the group mean receive positive advantages, while below-average outputs get negative advantages. Notably, this removes the need to train a \textit{critic} network to predict expected rewards, simplifying the algorithm and reducing variance by leveraging the sample-wise comparisons.

We optimize the policy $\pi_\theta$ by maximizing a clipped surrogate objective similar in form to PPO, but using the group-derived advantages $\bar{A}_{i,t}$ (for output $i$ at token $t$). The \textit{GRPO objective} can be written as:

{\small
\begin{align*}
&J_{\text{GRPO}}(\theta) = \\
&\mathbb{E}_{q, \{o_i\} \sim \pi_{\text{old}}(O|q)} \bigg\{ 
    \frac{1}{G} \sum_{i=1}^{G} \frac{1}{|o_i|} \sum_{t=1}^{|o_i|} \big\{ 
\min [
    \frac{\pi_\theta(o_{i,t} | q, o_{i,<t})}{\pi_{\text{old}}(o_{i,t} | q, o_{i,<t})} \bar{A}_{i,t},\;
    \text{clip} \left( 
        \frac{\pi_\theta(o_{i,t} | q, o_{i,<t})}{\pi_{\text{old}}(o_{i,t} | q, o_{i,<t})}, 
        1 - \epsilon, 1 + \epsilon 
    \right) \bar{A}_{i,t}
] \\
&   - \beta \mathcal{D}_{\text{KL}}(\pi_\theta \| \pi_{\text{ref}}) 
\big\} \bigg\}.
\end{align*}
}

Here $\epsilon$ and $\beta$ are hyperparameters. The inner summation averages the loss over all tokens in each output $o_i$, and the outer summation averages over the $G$ outputs in the group. GRPO also includes an explicit KL-divergence regularization term weighted by $\beta$, which keeps the updated policy from drifting too far from a reference policy (typically the pre-trained or SFT model).

The policy gradient for GRPO follows from this objective, and the update encourages the model to generate outputs that outperform its peer group's average, and discourages those that underperform, within the safe update step size enforced by the clipping.

\subsection*{Our GRPO Setup}

We implemented GRPO via the \textit{Hugging Face TRL} library. We set the group size $G = 4$, meaning for each question, the model (our policy) generates 4 candidate answers at each update step. These 4 answers are evaluated and used to compute advantages for policy gradient updates.

\subsection*{Reward Design}

Designing an appropriate reward function is crucial. We define a reward that captures two objectives: (1) Correctness of the answer, and (2) Proper format with reasoning. The reward for a generated answer is the sum of two components:

\begin{itemize}
    \item \textbf{Answer Correctness Reward:} If the model’s chosen option (the content inside \texttt{<answer>...</answer>} tags) matches the ground truth correct option, we give a reward of +2; otherwise 0. This strongly incentivizes selecting the right answer.
    \item \textbf{Format Compliance Reward:} If the model’s output strictly adheres to the expected format -- i.e., it includes a \texttt{<explanation>} section followed by an \texttt{<answer>} section, with no extraneous text -- then we give a reward of +1; otherwise 0. We implemented this by using a regex pattern to check the output matches the exact format template.
\end{itemize}

Thus, the maximum reward an output can receive is 3 (2+1) if it is perfectly formatted and the answer is correct. Partial reward is given for getting the format right even if the answer is wrong (to ensure the model doesn’t stop showing its reasoning) or vice versa (conceivably a correct guess but wrong format, though that’s rare when format is enforced during prompting). In practice, all high-reward outputs will contain a reasoning followed by the correct answer.

\subsection*{Training Procedure}

We initialized the policy for RL with the existing instruction model (which is our base model). For each training question, at each iteration, we sample 4 outputs from the current policy (with default temperature 0.7, to ensure some diversity). We included a KL-divergence penalty ($\beta = 0.005$) to keep the new policy from drifting too far from the initial policy distribution. The optimization was done with AdamW and the learning rate was 1e-5. We ran the reinforcement fine-tuning for 2 {epochs} over the training set. We set this set of parameters for training all models in {Figure~\ref{fig2}} to keep the results consistent. Our training infrastructure was a cluster with 8 H100 GPUs.

\section{Expanded Empirical Results by Category and Difficulty }
\label{app:category_diff}

\subsection{Category and Difficulty Assignment Pipeline}

To enable a more structured understanding and evaluation of the dataset, we systematically annotated each question with two additional attributes: a \textbf{question category} (reflecting its thematic focus) and a \textbf{difficulty level} (reflecting the cognitive complexity required to answer it). This classification process was designed to capture the rich diversity of scientific questions encountered in the dataset while also enabling stratified analyses along meaningful axes.

\subsubsection*{Category Assignment Process}

The first dimension of annotation involved categorizing questions into one of seven well-defined types: \textit{Validation, Troubleshooting \& Optimization}, \textit{Cloning \& Plasmid Construction}, \textit{Gene-editing Enzyme Selection}, \textit{GuideRNA Design}, \textit{CRISPR Screening \& Library Workflows}, \textit{Gene-editing Delivery Methods}, and \textit{Practical Considerations \& Lab Logistics}. These categories were carefully designed based on a detailed survey of the corpus and informed by domain expertise, ensuring coverage across different aspects of scientific inquiry.

To automate the classification, we implemented a keyword- and pattern-matching pipeline. Specifically, the categorization script parses each question’s text, searching for the occurrence of pre-specified sets of keywords or indicative phrases associated with each category. For example, mentions of \textit{experimental controls}, \textit{replicates}, or \textit{validation metrics} are mapped to the \textit{Validation, Troubleshooting \& Optimization} category. Where applicable, semantic context is also incorporated to disambiguate terms that may otherwise overlap across categories. The implementation prioritizes robustness by ensuring that the classification rules are mutually exclusive where possible, and fallback rules are applied if a question does not match high-confidence patterns.

\subsubsection*{Difficulty Level Assignment Process}

Besides, we assigned a difficulty rating to each question, choosing among three levels: \textit{Easy}, \textit{Medium}, and \textit{Hard}. The difficulty level reflects the anticipated cognitive effort required to generate a scientifically sound and complete response. Several heuristics guide this assignment:

\begin{itemize}
    \item Questions seeking factual recall, standard procedures, or straightforward clarifications were labeled \textit{Easy}.
    \item Questions requiring moderate inference, multi-step reasoning, or troubleshooting strategies were labeled \textit{Medium}.
    \item Questions demanding deep domain expertise, critical thinking about non-standard scenarios, or novel experimental design considerations were labeled \textit{Hard}.
\end{itemize}

The difficulty assignment similarly uses keyword signals, but it also incorporates structural indicators such as the length of the question, the presence of conditional clauses (``if...then'', ``how would it change if...''), and linguistic markers of uncertainty or hypothesis generation (``could it be that...'', ``what would happen if...''). This allows the system to infer the implicit reasoning demands behind each question.

\begin{figure}[htbp]
  \centering
  \includegraphics[width=1\textwidth]{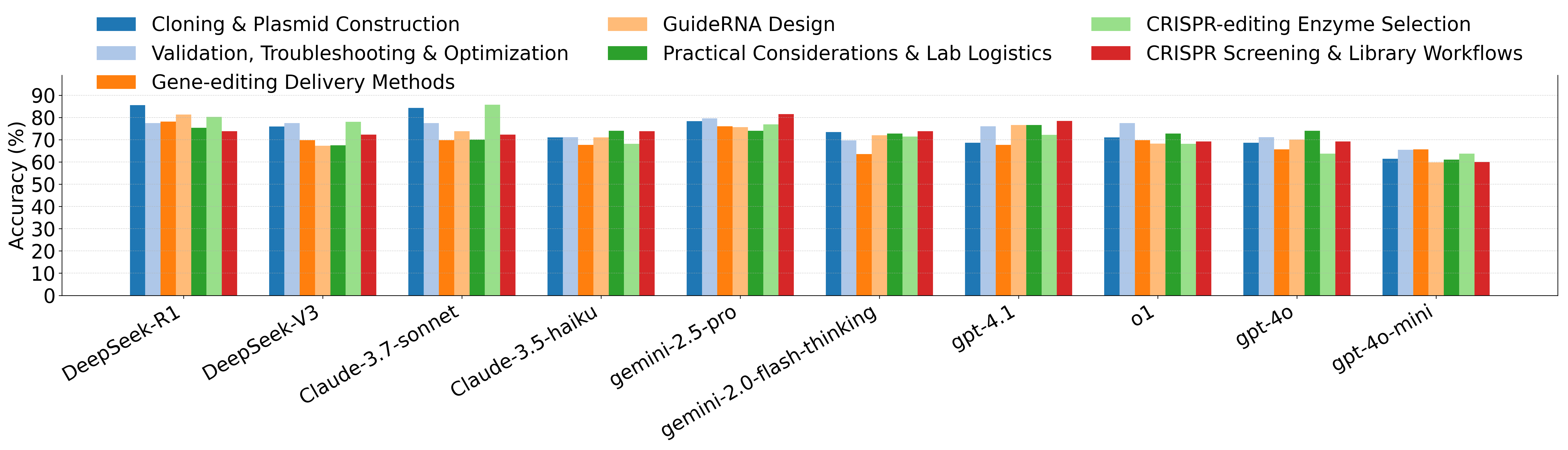}
  \caption{Model accuracy across seven genome categories for commercial LLMs.}
  \label{fig8}
\end{figure}

\begin{figure}[htbp]
  \centering
  \includegraphics[width=1\textwidth]{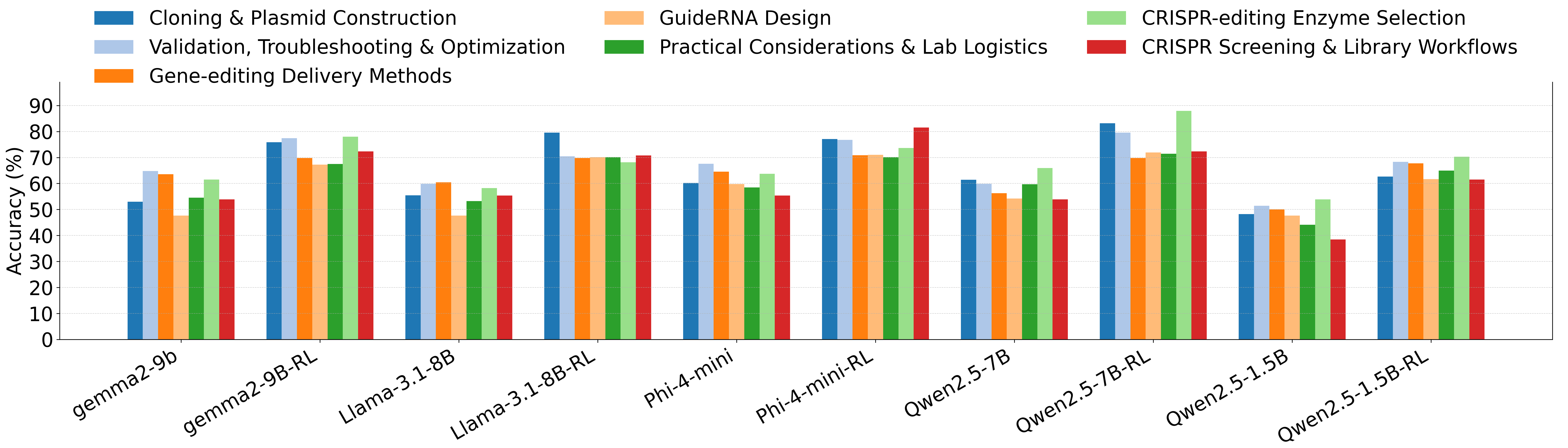}
  \caption{Model accuracy across seven genome categories for base and RL-fine-tuned models.}
  \label{fig9}
\end{figure}

\begin{figure}[htbp]
  \hspace{-2em}
  \includegraphics[width=1.1\textwidth]{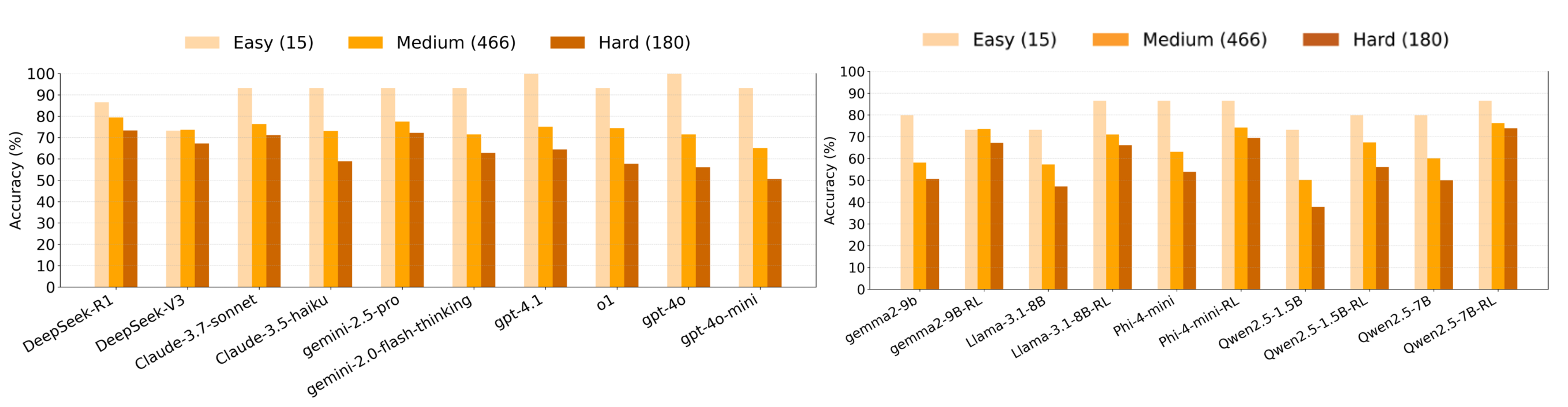}
  \caption{Accuracy of models across difficulty levels. Left Commercial models in the left and base and RL-fine-tuned models in the right.}
  \label{fig10}
\end{figure}

\subsection{Expanded Empirical Results According to Category and Difficulty }

By enriching each question with structured metadata regarding its category and difficulty, we enable a more nuanced analysis of model performance and human-AI comparisons. Stratifying evaluation results by question category allows us to investigate model strengths and weaknesses across different scientific competencies, while difficulty stratification provides insight into models’ robustness as question complexity increases.

First, category-stratified evaluation across reinforcement fine-tuned models shows that performance varies substantially across scientific domains (Figure~\ref{fig8},\ref{fig9}). For instance, \texttt{Llama-3.1-8B-RL} and \texttt{Qwen2.5-7B-RL} achieve higher accuracy on \textit{Cloning \& Plasmid Construction} but underperform on \textit{CRISPR Screening \& Library Workflows}. Meanwhile, \texttt{Phi-4-mini-RL} performs the best for \textit{CRISPR Screening \& Library Workflows} and underperforms on \textit{Cloning \& Plasmid Construction}. This indicates domain-specific gaps in model reasoning that would be invisible without categorical breakdowns.

Second, difficulty-stratified results on commercial models (e.g., \texttt{DeepSeek-R1}, \texttt{Claude}, \texttt{Gemini}, \texttt{GPT-4} variants) reveal a consistent trend: while models perform well on \textit{Easy} questions, their accuracy drops sharply for \textit{Hard} questions, especially in smaller models like \texttt{GPT-4o-mini} (see Figure~\ref{fig10}). This suggests that current models struggle with more complex biological reasoning, underscoring the importance of incorporating question difficulty into evaluation.

Third, reinforcement learning (RL)-fine-tuned models demonstrate clear performance gains across most categories compared to their instruction counterparts (base models), suggesting that targeted fine-tuning strategies can systematically strengthen model weaknesses identified through categorical and difficulty analyses. However, the magnitude of improvement is not uniform, suggesting that further targeted fine-tuning may be needed to close residual gaps in specific scientific subfields.

\section{Extended Ablation Studies of Fine-Tuned Models}

\subsection*{Supervised Fine-tuning in Ablation Figure~\ref{fig3}}

Supervised fine-tuning involves training a large language model (LLM) to perform well on tasks where it follows user instructions. It’s a process where the model is fine-tuned on labeled datasets, where each input corresponds to a specific desired output. The goal is to align the model's behavior with the human expert’s intent. Let the model be parameterized by $\theta$, and given an input question $x$, the model outputs a probability distribution $P_{\theta}(y|x)$ over the possible outputs $y$.

During supervised fine-tuning, the model is trained on pairs $(x_i, y_i)$, where $x_i$ is the \textit{question} in Genome-Bench and $y_i$ is the \textit{answer} (has format like \texttt{<explanation>...</explanation><answer>...</answer>}).

The goal of SFT is to minimize the difference between the model's predicted output and the true output (ground truth) for a given instruction. The standard loss function used is the \textit{cross-entropy loss}, which measures how well the predicted probability distribution $P_{\theta}(y_i | x_i)$ aligns with the actual distribution. The cross-entropy loss for a single instruction-output pair $(x_i, y_i)$ is defined as:

\[
L(\theta; x_i, y_i) = -\sum_{t=1}^{T} \log P_{\theta}(y_{i,t} \mid x_i, y_{i,<t}),
\]

where $y_{i,t}$ is the token at position $t$ in the output sequence $y_i$, $T$ is the length of the output sequence, and $P_{\theta}(y_{i,t} \mid x_i, y_{i,<t})$ is the probability assigned by the model to the token $y_{i,t}$ conditioned on the input $x_i$ and all previously generated tokens $y_{i,<t}$.

For the Genome-Bench dataset pairs $\{(x_i, y_i)\}_{i=1}^N$, the total loss is:

\[
L(\theta) = \frac{1}{N} \sum_{i=1}^{N} L(\theta; x_i, y_i).
\]

This loss function encourages the model to assign higher probabilities to correct outputs (i.e., $y_i$) for a given input instruction $x_i$.

To minimize the loss, the parameter update rule at step $t$ is given by:

\[
\theta_{t+1} = \theta_t - \eta \nabla_{\theta} L(\theta_t),
\]

where the learning rate $\eta$ is a hyperparameter.

\begin{figure}[t]
  \centering

  \begin{minipage}[b]{0.45\textwidth}
    \centering
    \includegraphics[width=\textwidth]{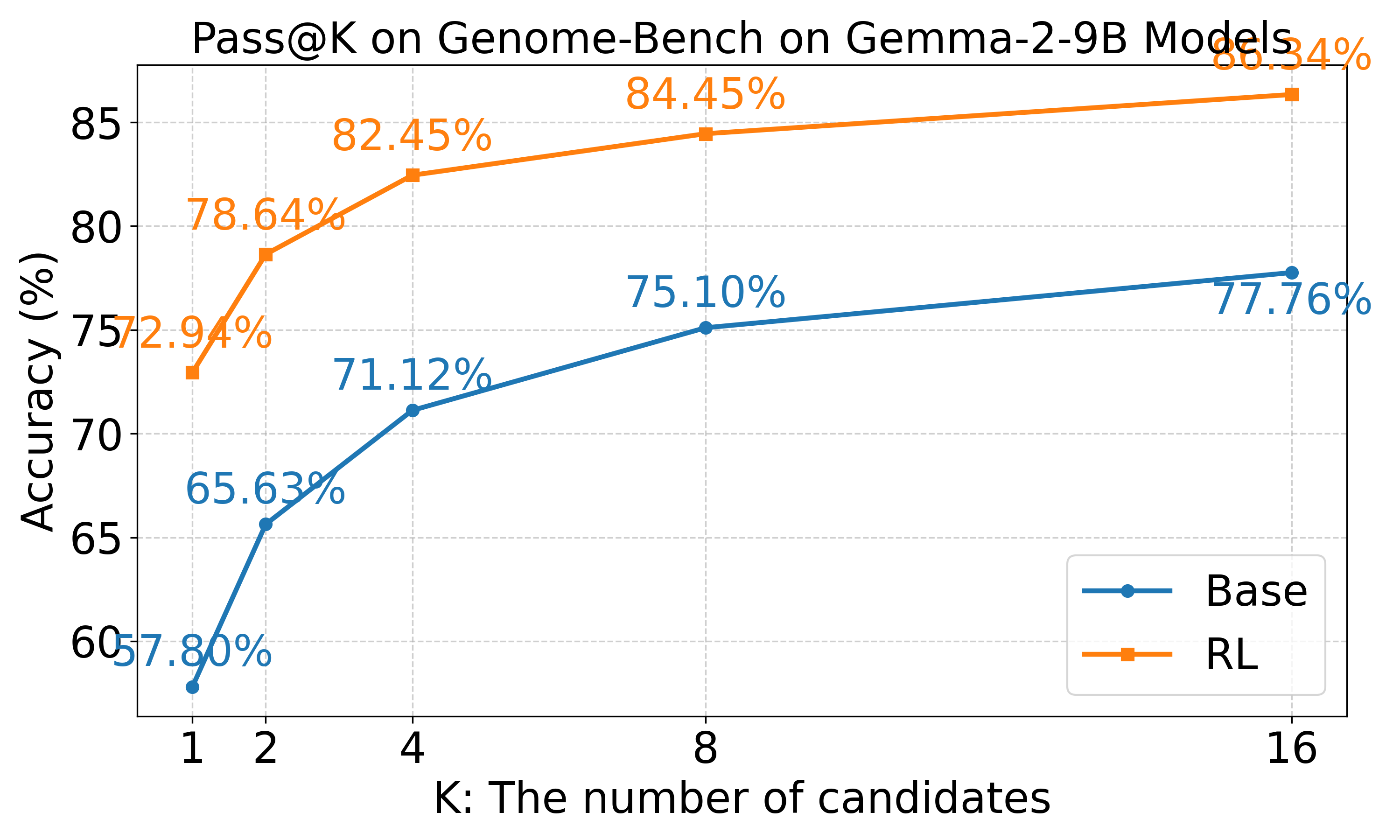}
  \end{minipage}
  \hfill
  \begin{minipage}[b]{0.45\textwidth}
    \centering
    \includegraphics[width=\textwidth]{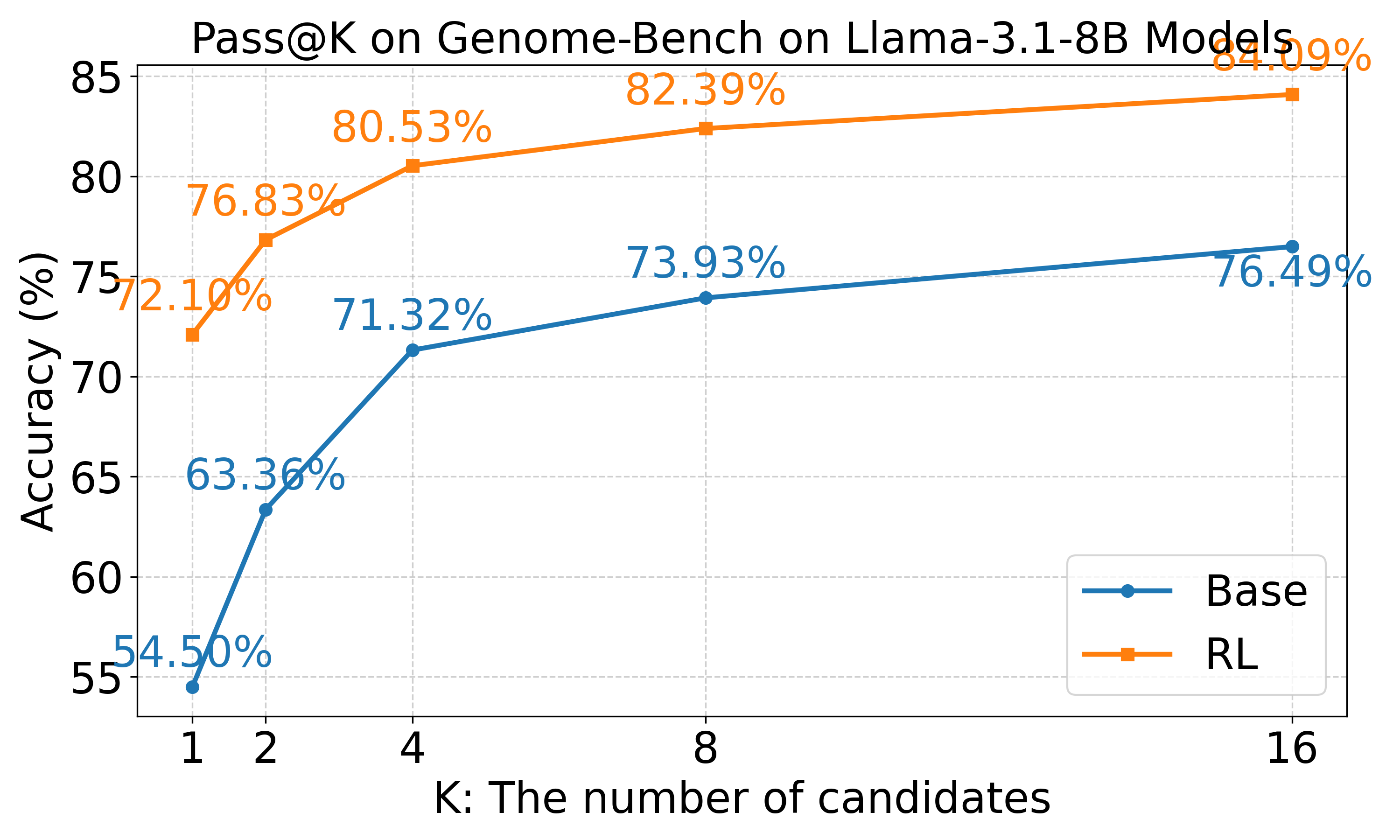}
  \end{minipage}

  \begin{minipage}[b]{0.45\textwidth}
    \centering
    \includegraphics[width=\textwidth]{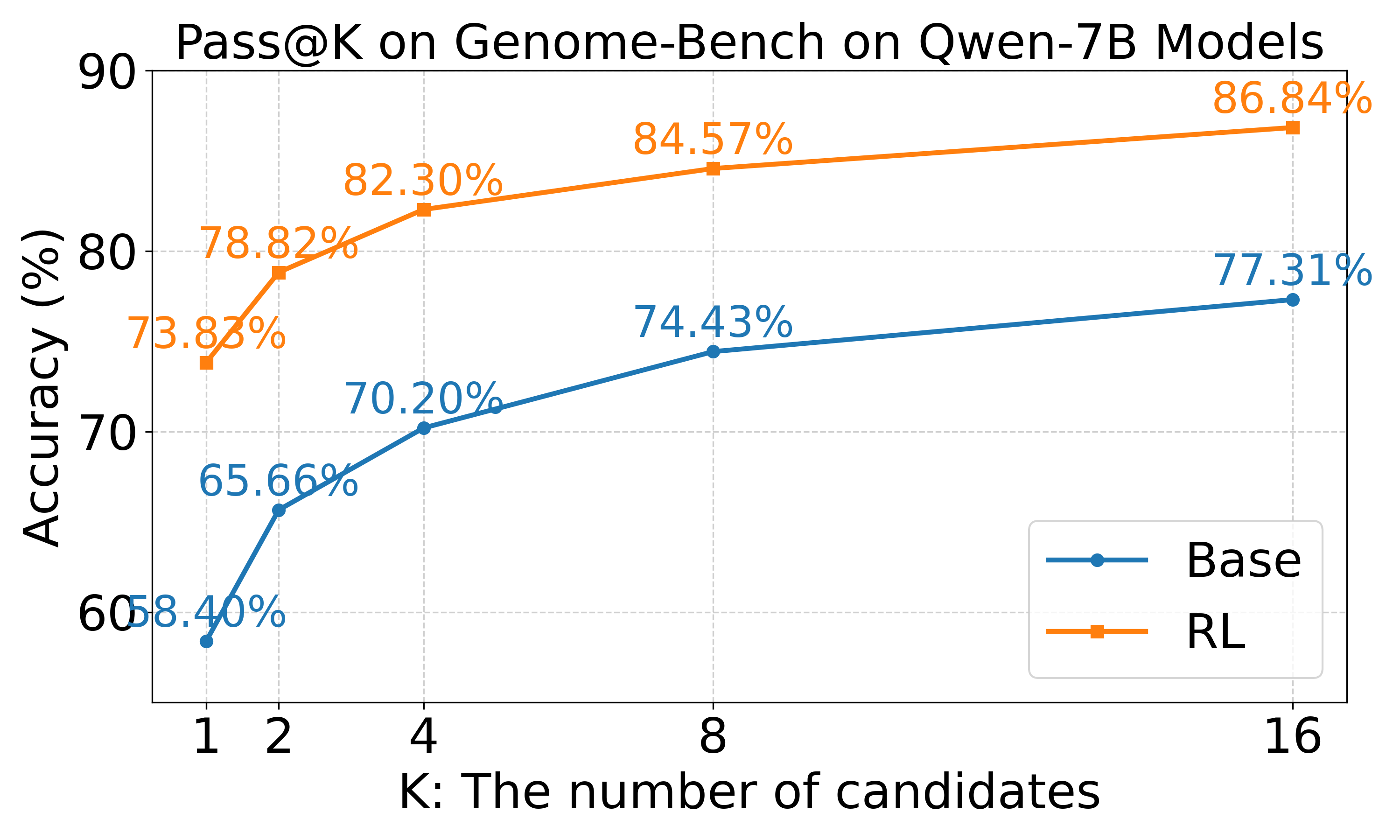}
  \end{minipage}
 \hfill
  \begin{minipage}[b]{0.45\textwidth}
    \centering
    \includegraphics[width=\textwidth]{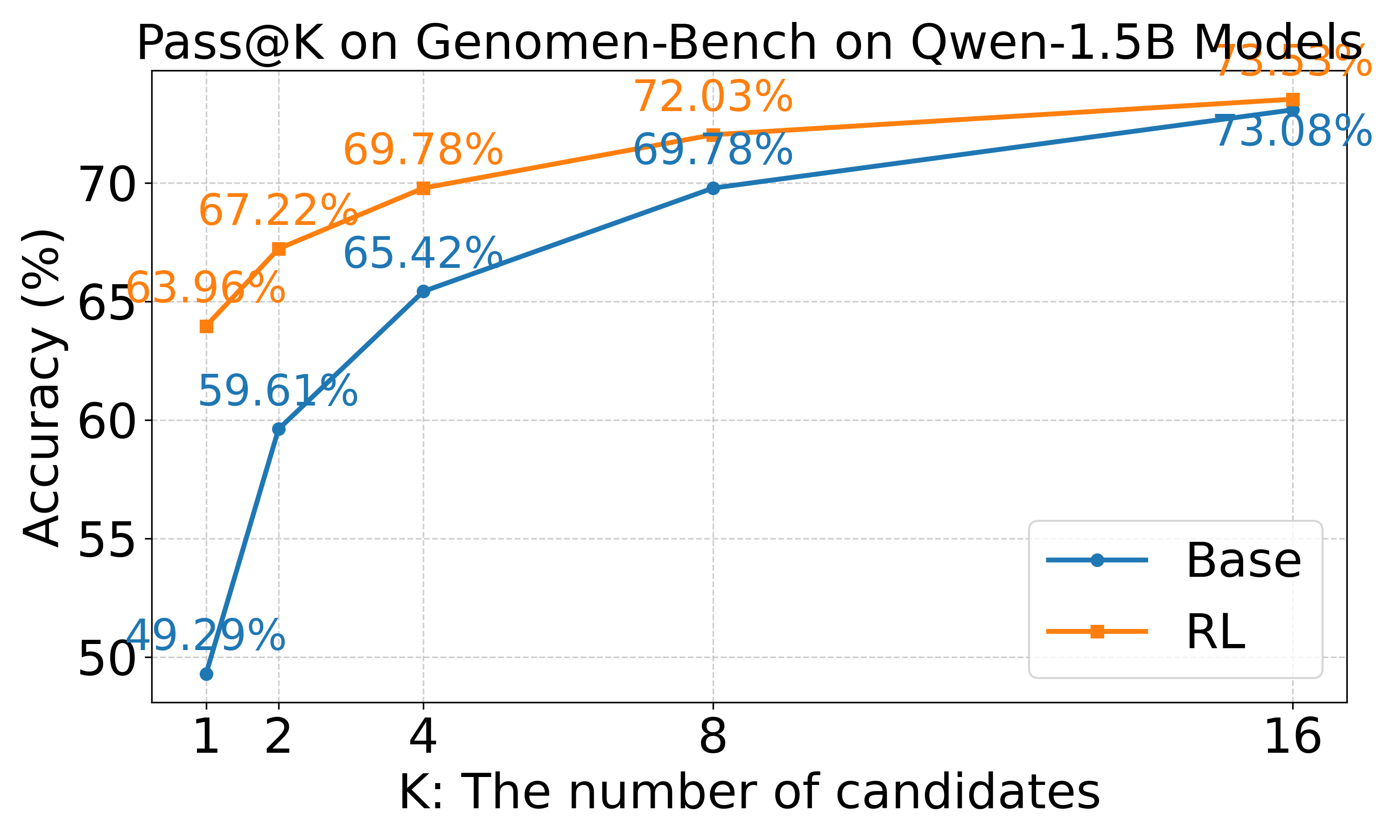}
  \end{minipage}

  \caption{\emph{Pass@K evaluation for RL vs Base models.} All models are evaluated for $K{=}1,2,4,8,16$. The Pass@K criterion considers a question correctly answered if at least one of the $K$ independently sampled outputs contains the correct answer.}
  \label{fig:passatk}
\end{figure}

\subsection*{Pass@K Experiment in Figure~\ref{fig:passatk}.} For both \texttt{Gemma-2-9B} and \texttt{Llama-3.1-8B}, RL fine-tuning leads to consistent and substantial improvements over their base counterparts as $K$ increases, with the performance gap remaining wide even at larger $K$ values. This pattern suggests that RL enhances not only output diversity but also imparts stronger reasoning capabilities to these larger models. For \texttt{Qwen2.5-7B}, RL also consistently outperforms the base model, maintaining a clear margin as $K$ increases, suggesting acquisition of new reasoning patterns through fine-tuning. In contrast, for the smaller \texttt{Qwen2.5-1.5B} model, the gap between the RL and base versions narrows as $K$ increases, implying that RL primarily enhances output robustness rather than introducing fundamentally new capabilities, likely due to the limited capacity of the smaller model.

\begin{figure}[ht]
  \centering
  \begin{minipage}[b]{0.65\textwidth}
    \centering
    \includegraphics[width=\textwidth]{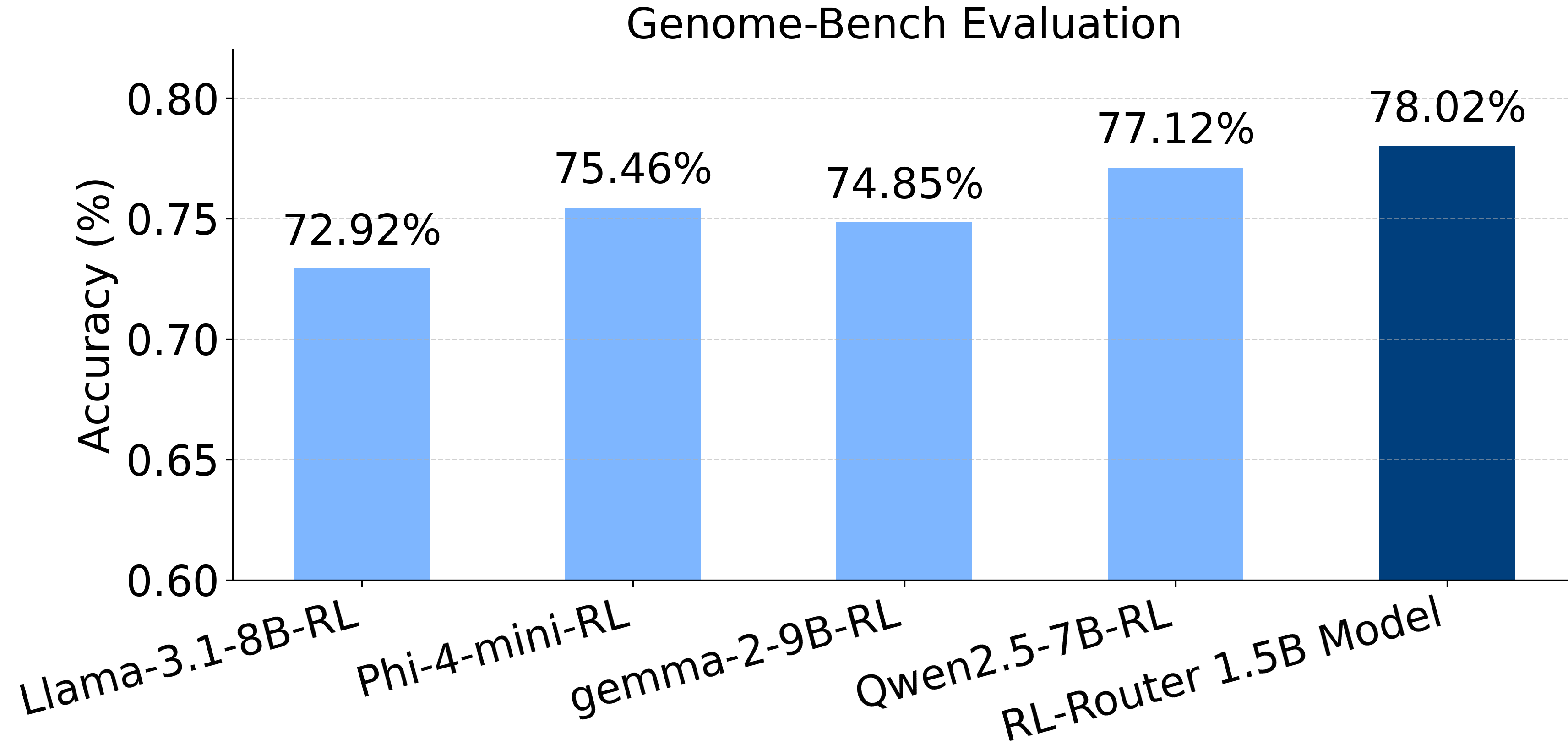}
  \end{minipage}
  \hfill
  \begin{minipage}[b]{0.34\textwidth}
    \centering
    \includegraphics[width=\textwidth]{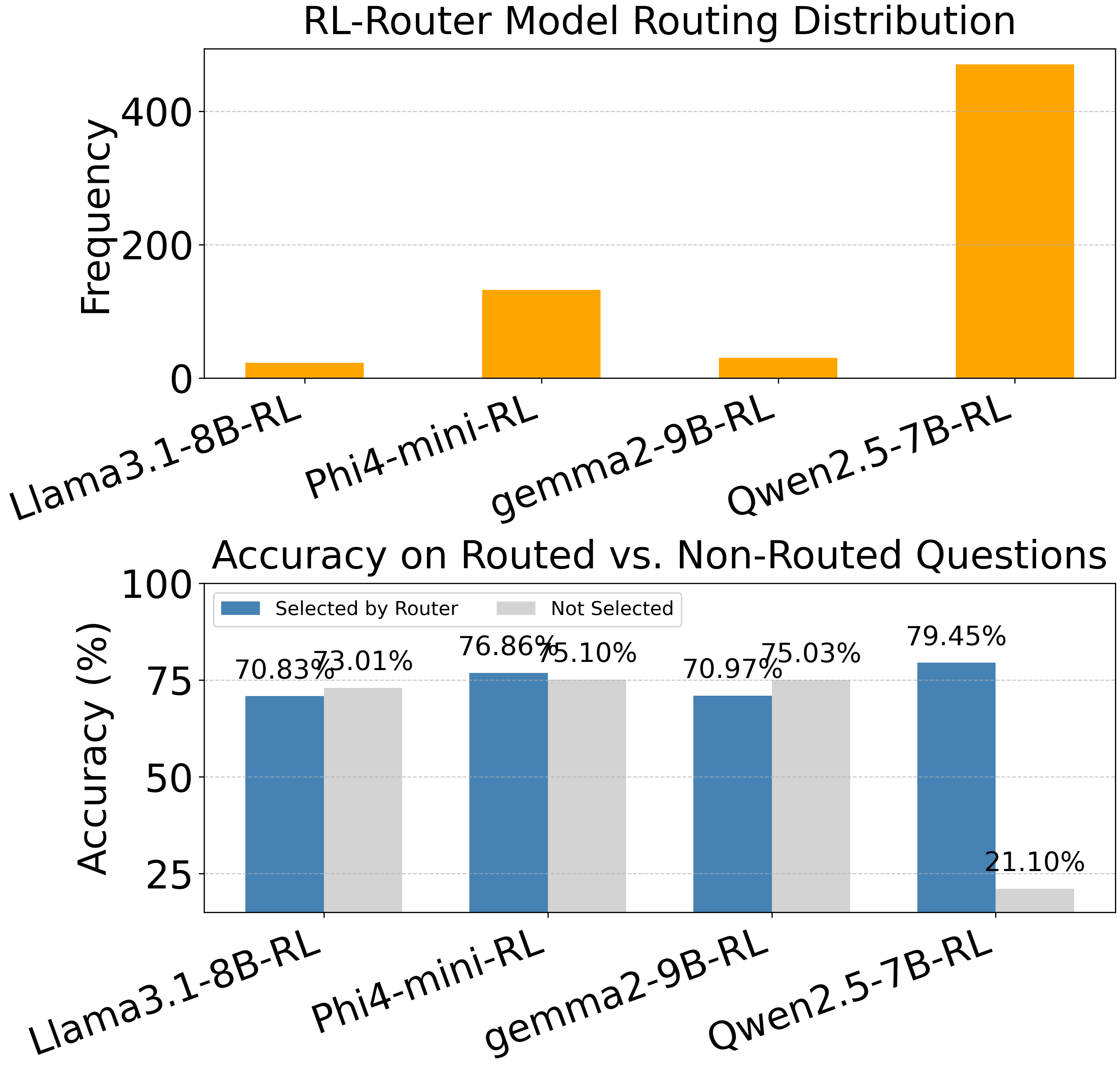}
  \end{minipage}
  \caption{\emph{Performance of 1.5B RL-Router model.} The backbone model used for training is \texttt{Qwen2.5-1.5B-Instruct} model. }
  \label{fig:rl_router_1.5B}
\end{figure}

\begin{figure}[ht]
  \centering
  \begin{minipage}[b]{0.45\textwidth}
    \centering
    \includegraphics[width=\textwidth]{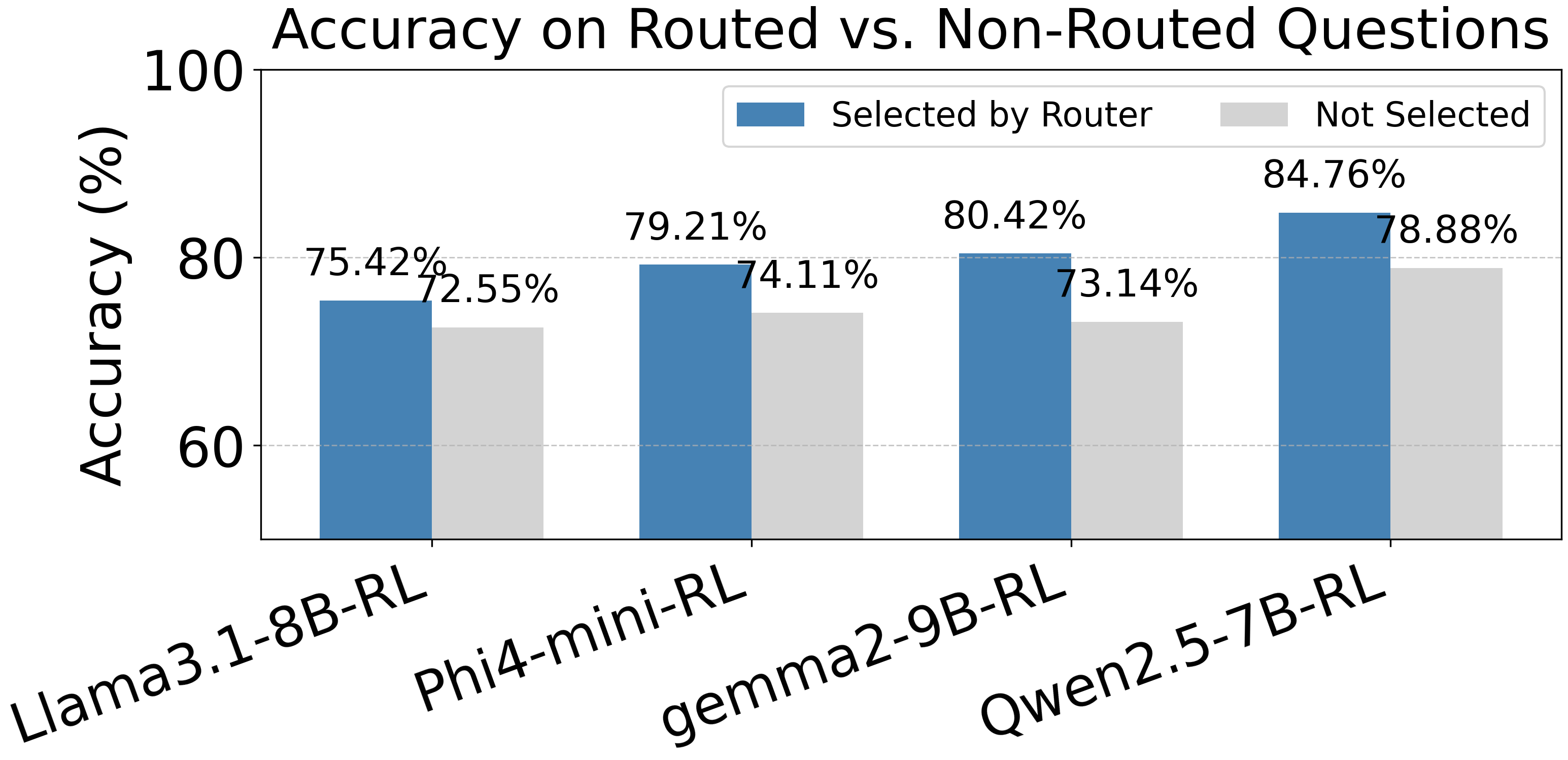}
  \end{minipage}
  \hfill
  \begin{minipage}[b]{0.45\textwidth}
    \centering
    \includegraphics[width=\textwidth]{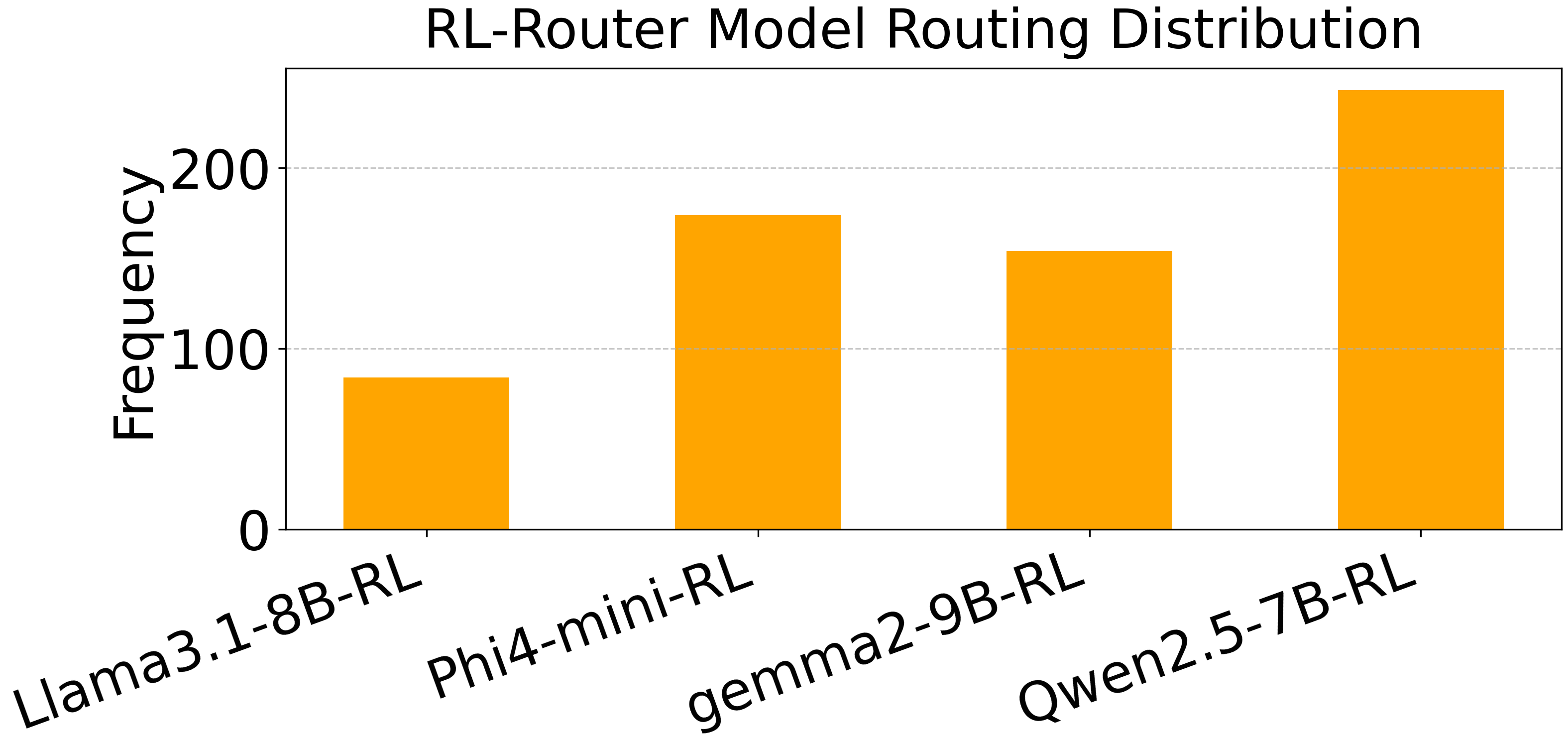}
  \end{minipage}
  \caption{Breakdown accuracy and routing distribution for 7B RL-Router model.}
  \label{fig:rl_router_7B_rl}
\end{figure}

\begin{figure}[ht]
  \centering
  \begin{minipage}[b]{0.45\textwidth}
    \centering
    \includegraphics[width=\textwidth]{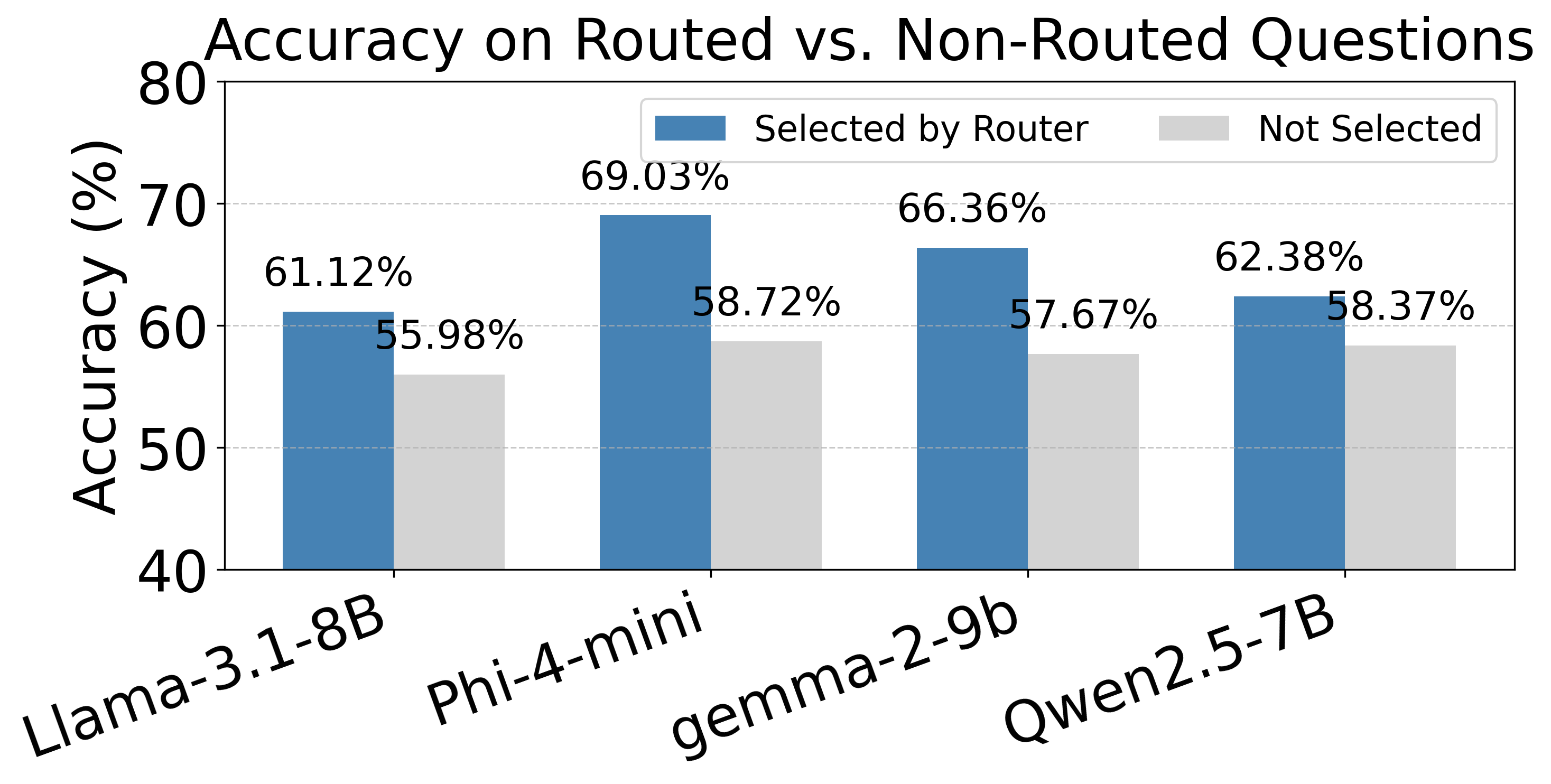}
  \end{minipage}
  \hfill
  \begin{minipage}[b]{0.45\textwidth}
    \centering
    \includegraphics[width=\textwidth]{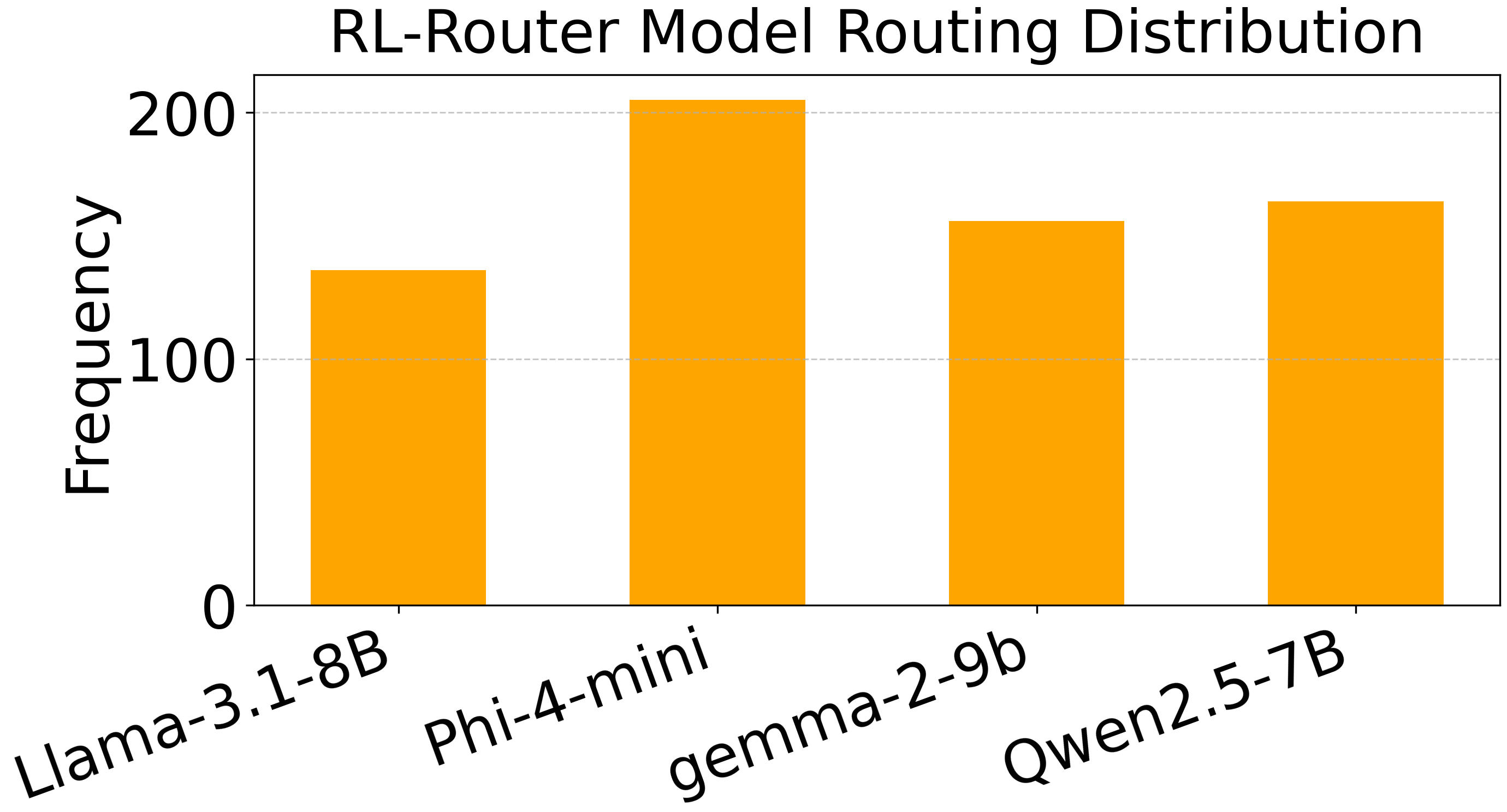}
  \end{minipage}
  \caption{Breakdown accuracy and routing distribution of 7B RL-Router model for Base Models.}
  \label{fig:rl_router_7B_base}
\end{figure}

\section{Additional Details on Router Models}
\label{app_sec:router}

\textbf{Objective.} Learn a policy (router) that selects the best expert model among a set of four RL-tuned LMs for each input question from Genome-Bench.

\textbf{Task Formulation.} Single-choice multi-class selection: for each question, the router outputs a choice \{1,2,3,4\} corresponding to one of the four RL experts.

\textbf{Reward Design.} The router receives: \textbf{+1} reward if the selected model answers correctly; \textbf{--1} reward if the selected model answers incorrectly.

\textbf{Backbone Model.} Open-source language model \textit{Qwen2.5-7B-Instruct} adapted as a router.

\textbf{Modified RL Trainer.} A custom subclass \texttt{PreGeneratedGRPOTrainer} built on \texttt{GRPOTrainer} (from Hugging Face TRL library), designed to:
\begin{itemize}
    \item Use pre-generated candidate completions instead of on-the-fly sampling (for computation reason).
    \item Randomly sample one pre-generated choice per training iteration.
    \item Compute rewards based on correctness of selected model’s response.
\end{itemize}

\textbf{Training Configuration used in Figure~\ref{fig4}.} Batch size per device = 8, Group size = 8, Optimizer = \texttt{adamw}, Learning rate = 1e-5, epoch = 2, warmup ratio = 0.1, weight decay = 0.1, bfloat16 training.

\section{More Details of Human Evaluations and Model Reasoning in Figure~\ref{fig5}}

\textbf{Objective - Human Evaluation.} The goal of this evaluation is to benchmark the model’s performance not only based on answer correctness, but also on the quality of reasoning, using human experts as a reference standard.

\textbf{Task Design.} We randomly selected 35 questions from the Genome-Bench dataset with difficulty level \emph{Medium} or \emph{Hard} for detailed evaluation. All answer labels were removed to ensure unbiased responses. Three genome engineering experts were asked to independently complete the 35 questions. Each expert was instructed to provide their responses in a structured format:

\begin{quote}
\texttt{<explanation> Expert Reasoning </explanation> <answer> Expert Choice </answer>}
\end{quote}

The first field captured the expert’s step-by-step reasoning, while the second field specified their final selected option.

\textbf{Model Evaluation.} The same 35 questions were provided to three models:

\begin{itemize}
    \item \texttt{Qwen2.5-7B-RL} (the best-performing single RL-tuned model),
    \item \texttt{DeepSeek-R1} (the best-performing commercial model),
    \item \texttt{Router Model} (trained via reinforcement routing).
\end{itemize}

Each model answered the 35 questions independently using its inference pipeline, with outputs captured in the same format (reasoning + selected choice) for three times.

\textbf{Human-Model Comparison.} As shown in Figure~\ref{fig5}, human experts achieved an average accuracy of 73.3\% on these questions, setting a strong performance upper bound. Among the models, the RL-Router model achieved the highest accuracy (67.6\%), followed by \texttt{DeepSeek-R1} (65.7\%) and \texttt{Qwen2.5-7B-RL} (61.9\%). In addition to numerical accuracy, qualitative analysis of model explanations was performed. Several examples are shown in the figure:

\begin{itemize}
    \item \textbf{In cases with Correct Reasoning}, the model successfully selected the right answer and provided an explanation consistent with expert logic.
    \item \textbf{In Partially Wrong Reasoning} cases, the model identified the correct answer but introduced inaccuracies or incomplete logic in the explanation.
    \item \textbf{In Completely Wrong Reasoning} cases, even when selecting the correct answer, the model’s explanation diverged significantly from the scientific reasoning expected.
\end{itemize}

\section{Examples Correctly Answered by the RL Model but Missed by the Base Model}\label{app:right-wrong}

\begin{methodbox}{Example of Qwen2.5-7B Response: ID 7}
\vspace{0.1cm}
\textbf{Question:} 
How should duplicated gRNA sequences that appear to target the same gene family in the GeCKO V2 library be handled during analysis? 

\textbf{Please choose one of the following options:} 

a. Remove all duplicate guides from the analysis to ensure specificity. 

b. Use only one representative guide from each duplicated set for analysis.

c. Create a unique list of guides but keep track of guides that target more than one gene.

d. Ignore the duplicates as they do not affect the overall results.

e. Assign duplicates to a separate analysis group for further investigation.

\vspace{0.1cm}
\textbf{Correct Answer:} c. 
\vspace{0.1cm}

\textbf{Before RL:} <explanation>When analyzing the GeCKO V2 library, it's important to maintain specificity and accuracy. Keeping only one representative guide from each duplicated set ensures that the analysis is not biased and remains focused on the intended targets.</explanation> <answer>b</answer>.
\vspace{0.1cm}

\textbf{After RL:} <explanation>When analyzing the GeCKO V2 library, it's important to ensure that each gRNA sequence is unique to avoid redundancy and ensure specificity. However, if duplicates appear to target the same gene family, it's crucial to manage them properly to avoid false positives or negatives.</explanation> <answer>c</answer>.
\end{methodbox}

\begin{methodbox}{Example of Qwen2.5-7B Response: ID 23}
\vspace{0.1cm}
\textbf{Question:} 
I've been having trouble with some gene IDs not working in the CRISPR analysis tool you shared with me. Could you help me figure out how to solve the issue with IDs like NM\_004036.3 and NM\_015270.3? 

\textbf{Please choose one of the following options:} 

a. Use a manual annotation approach to identify correct gene ID. 

b. Use a different software tool that accepts these gene IDs. 

c. Contact the software support team for troubleshooting.

d. Type the gene name like `ADCY3' instead of the gene ID. 

e. Convert the gene IDs into protein sequences before analysis.

\vspace{0.1cm}
\textbf{Correct Answer:} d. 
\vspace{0.1cm}

\textbf{Before RL:} <explanation>The issue with non-working gene IDs like NM\_004036.3 and NM\_015270.3 could be due to compatibility or recognition by the analysis tool. Contacting the software support team for troubleshooting would be the most direct way to resolve this issue.</explanation> <answer>c</answer>.
\vspace{0.1cm}

\textbf{After RL:} <explanation>When encountering issues with specific gene IDs in CRISPR analysis tools, it's often helpful to ensure that the IDs are recognized. For gene IDs like NM\_004036.3 and NM\_015270.3, which are mRNA IDs, the tool might require the gene symbol instead.</explanation> <answer>d</answer>.

\end{methodbox}

\begin{methodbox}{Example of Qwen2.5-7B Response: ID 27}
\vspace{0.15cm}
\textbf{Question:} 
Is it possible that the oligonucleotides I'm using are somehow affecting the enzymatic reactions in my genetic engineering work? 

\vspace{0.15cm}
\textbf{Please choose one of the following options:} 

\vspace{0.15cm}
a. No, oligos do not typically interfere with enzymatic reactions in genetic engineering.

\vspace{0.15cm}
b. Yes, oligos can interfere by increasing the likelihood of off-target effects in CRISPR experiments. 

\vspace{0.15cm}
c. No, oligos are specifically designed to avoid interference in genetic engineering protocols.

\vspace{0.15cm}
d. Yes, using a simultaneous digestion-ligation protocol can minimize self-ligation and improve results.

\vspace{0.15cm}
e. Yes, oligos may cause unexpected outcomes due to non-specific binding.

\vspace{0.15cm}
\textbf{Correct Answer:} d. 

\vspace{0.15cm}
\textbf{Before RL:} <explanation>Oligonucleotides, while crucial for genetic engineering, can indeed interfere with enzymatic reactions. They can cause non-specific binding, leading to unexpected outcomes in genetic engineering experiments.</explanation> <answer>e</answer>.

\vspace{0.15cm}
\textbf{After RL:} explanation>Based on the context, the most relevant and practical answer is that using a simultaneous digestion-ligation protocol can help manage the self-ligation of oligonucleotides, thereby improving the results of genetic engineering work.</explanation> <answer>d</answer>.

\end{methodbox}

\begin{methodbox}{Example of Qwen2.5-7B Response: ID 29}
\vspace{0.15cm}
\textbf{Question:} 
In the context of a genome engineering discussion, how would you specifically go about integrating genes from different species to potentially create superhuman abilities, and which genes would you choose? 

\vspace{0.15cm}
\textbf{Please choose one of the following options:} 

\vspace{0.15cm}
a. Genes from a spider integrated into humans for enhanced agility and strength.

\vspace{0.15cm}
b. Genes from a fish integrated into birds for superior flight ability. 

\vspace{0.15cm}
c. Genes from a plant integrated into animals for photosynthesis capability.

\vspace{0.15cm}
d. Genes from a bear integrated into a whale for hibernation in aquatic environments. 

\vspace{0.15cm}
e. Genes from a bacteria integrated into humans for antibiotic resistance.

\vspace{0.15cm}
\textbf{Correct Answer:} a.

\vspace{0.15cm}
\textbf{Before RL:} <explanation>While the idea of integrating genes from different species to create superhuman abilities is speculative and involves complex ethical and biological considerations, the most scientifically grounded and plausible option would be integrating genes for specific traits that are well-documented and have a clear function. Antibiotic resistance genes from bacteria, while not directly related to superhuman abilities, represent a well-understood genetic trait with significant biomedical relevance.</explanation> <answer>e</answer>.

\vspace{0.15cm}
\textbf{After RL:} <explanation>The question is about integrating genes from different species to potentially create superhuman abilities. Spider genes (option a) are a well-known example used in science fiction and popular culture for enhancing human agility and strength.</explanation> <answer>a</answer>.

\end{methodbox}

\begin{methodbox}{Example of Qwen2.5-7B Response: ID 36}
\vspace{0.2cm}
\textbf{Question:} 
Could you offer some advice on using CRISPR technology in drosophila, specifically regarding whether the homology arms need to flank the target site and if selection markers should be included in donor plasmids? 

\vspace{0.2cm}
\textbf{Please choose one of the following options:}

\vspace{0.2cm}
a. No, if I design homology arm just a few bases away from the target site so it will save the extra effort to mutate the PAM site.

\vspace{0.2cm}
b. No, it is not necessary if selection markers are used.

\vspace{0.2cm}
c. Yes, this ensures precise integration at the target site. 

\vspace{0.2cm}
d. Only if the target site is within a coding region.

\vspace{0.2cm}
e. Yes, to facilitate the use of CRISPR technology effectively.

\vspace{0.2cm}
\textbf{Correct Answer:} a. 

\vspace{0.2cm}
\textbf{Before RL:} <explanation>For CRISPR in drosophila, it is generally recommended to include homology arms that flank the target site to ensure precise integration. While selection markers in donor plasmids can be useful, they do not replace the need for homology arms for precise integration.</explanation> <answer>c</answer>.

\vspace{0.2cm}
\textbf{After RL:} <explanation>In drosophila CRISPR applications, it is often sufficient to have homology arms a few bases away from the target site, as the PAM site does not need to be mutated. Selection markers in donor plasmids are also useful but not strictly necessary.</explanation> <answer>a</answer>.

\end{methodbox}

\begin{methodbox}{Example of Qwen2.5-7B Response: ID 58}
\vspace{0.2cm}
\textbf{Question:} 
Could you provide me with the complete sequence of the original LentiCRISPR used in the CRISPR/Cas9 system? 

\vspace{0.2cm}
\textbf{Please choose one of the following options:} 

\vspace{0.2cm}
a. The sequence is proprietary and not publicly available. 

\vspace{0.2cm}
b. You can find it on the official university website. 

\vspace{0.2cm}
c. It is only available to registered researchers.

\vspace{0.2cm}
d. It is available in the supplementary materials of the original research paper. 

\vspace{0.2cm}
e. Here you go! http://genome-engineering.org/gecko/?page\_id=34.

\vspace{0.2cm}
\textbf{Correct Answer:} e.

\vspace{0.2cm}
\textbf{Before RL:} <explanation>The sequence of the original LentiCRISPR-v2 was provided in the supplementary materials of the original research paper by DiCarlo et al. published in 2014.</explanation> <answer>d</answer>.

\vspace{0.2cm}
\textbf{After RL:} <explanation>The original LentiCRISPR plasmid sequence is publicly available through the provided link, which is a well-known resource for genome engineering tools.</explanation> <answer>e</answer>.

\end{methodbox}

\begin{methodbox}{Example of Qwen2.5-7B Response: ID 63}
\vspace{0.15cm}
\textbf{Question:} 
When you were working on modifying the genetic sequences using CRISPR, did you make any changes to the PAM site on your repair template? 

\vspace{0.15cm}
\textbf{Please choose one of the following options:} 

\vspace{0.15cm}
a. No, the PAM site is not related to repair template design.

\vspace{0.15cm}
b. Yes, modifying the PAM site helps to prevent re-cutting by the CRISPR system.

\vspace{0.15cm}
c. Yes, altering the PAM site is required to guide the Cas9 to the target site.

\vspace{0.15cm}
d. Yes, changing the PAM site improves the efficiency of non-homologous end joining.

\vspace{0.15cm}
e. No, changing the PAM site is not necessary for successful editing.

\vspace{0.15cm}
\textbf{Correct Answer:} b. 

\vspace{0.15cm}
\textbf{Before RL:} <explanation>The PAM site is a specific sequence to which Cas9 binds for gene editing. Modifying it is not typically necessary or beneficial for the repair template design.</explanation> <answer>a</answer>.

\vspace{0.15cm}
\textbf{After RL:} <explanation>Typically, when using CRISPR, one does not modify the PAM site on the repair template because the PAM (Protospacer Adjacent Motif) is a recognition site for the Cas9 enzyme, which is essential for target recognition and cleavage. Changing the PAM site could interfere with the Cas9's ability to bind and cleave the DNA at the desired site.</explanation> <answer>b</answer>.

\end{methodbox}

\begin{methodbox}{Example of Qwen2.5-7B Response: ID 71}
\vspace{0.15cm}
\textbf{Question:} 
Has anyone experimented with using Cas9 mRNA that includes a polyA signal instead of polyadenylating it in vitro, especially for effectiveness in mouse zygotes at the one-cell stage? 

\vspace{0.15cm}
\textbf{Please choose one of the following options:} 

\vspace{0.15cm}
a. A polyA signal is not recognized in mammalian systems

\vspace{0.15cm}
b. I know from the Xenopus field that a polyA signal is sufficient

\vspace{0.15cm}
c. In vitro polyadenylation is necessary for Cas9 mRNA stability 

\vspace{0.15cm}
d. Using a polyA signal instead of in vitro polyadenylation is ineffective in mouse zygotes

\vspace{0.15cm}
e. A polyA signal requires additional modifications to function in Xenopus

\vspace{0.15cm}
\textbf{Correct Answer:} b. 

\vspace{0.15cm}
\textbf{Before RL:} <explanation>The polyA signal itself is not sufficient for stability in mammalian systems, including mouse zygotes. In vitro polyadenylation typically ensures the necessary stability and function of Cas9 mRNA.</explanation> <answer>d</answer>.

\vspace{0.15cm}
\textbf{After RL:} <explanation>I have knowledge from the Xenopus field where a polyA signal is sufficient for mRNA stability and function. The same could potentially apply to mouse zygotes, though specific experiments would be needed to confirm.</explanation> <answer>b</answer>.

\end{methodbox}

\begin{methodbox}{Example of Qwen2.5-7B Response: ID 109}
\vspace{0.1cm}
\textbf{Question:} 
I'm new to using CRISPR and lentiviruses in my research and I have a question about planning my experiments. I currently have cells expressing the lentiCas9-blast construct and I'm using guide RNAs from the lentiGuide-Puro constructs to create knockouts. I want to complement my gene of interest and have expression vectors ready, but they all have Puro resistance. Since these vectors integrate, could you recommend another common selectable marker for the third vector, or should I screen for GFP or something similar to confirm transduction? 

\textbf{Please choose one of the following options:} 

a. Conduct an antibiotic titration experiment to identify the optimal concentration.

b. Use a different fluorescent marker like RFP.

c. Switch to a different antibiotic resistance such as Hygromycin. 

d. Use GFP selection if available.

e. None of the above.

\vspace{0.1cm}
\textbf{Correct Answer:} d. 
\vspace{0.1cm}

\textbf{Before RL:} <explanation>In your case, since you are already using Puro resistance for the lentiGuide-Puro constructs, you should switch to a different selectable marker to avoid selection pressure on the same resistance. Hygromycin is a common alternative that can be used for selection in this scenario.</explanation> <answer>c</answer>.
\vspace{0.1cm}

\textbf{After RL:} <explanation>Given that your cells already have Puro resistance and you are using it for the lentiGuide-Puro constructs, it would be best to use a different selectable marker for the third vector to avoid potential resistance issues or to screen for transduction efficiency. GFP could be used as a selectable marker to confirm transduction, as it is commonly used for this purpose.</explanation> <answer>d</answer>".

\end{methodbox}

\begin{methodbox}{Example of Qwen2.5-7B Response: ID 140}
\vspace{0.1cm}
\textbf{Question:} 
Could you clarify the design and suitability of the primers used in the second PCR for the GECKO library, particularly in relation to Illumina sequencing? 

\textbf{Please choose one of the following options:} 

a. The final amplicon should be longer than 1000 bp to ensure compatibility with Illumina sequencing.

b. Primers must always be synthesized with a GC content of 60\% for GECKO libraries.

c. It is important to use primers with staggered annealing temperatures to avoid non-specific binding.

d. Using primers that result in an amplicon of <800 bp is ideal for Illumina readout.

e. The primers should be designed with stagger on F and barcoding on R for optimal Illumina sequencing.

\vspace{0.1cm}
\textbf{Correct Answer:} e. 
\vspace{0.1cm}

\textbf{Before RL:} <explanation>The primers for the GECKO library should be designed with specific characteristics to ensure compatibility with Illumina sequencing. Staggered annealing temperatures help avoid non-specific binding, which is crucial for accurate sequencing results.</explanation> <answer>c</answer>.
\vspace{0.1cm}

\textbf{After RL:} <explanation>The primers for the second PCR in the GECKO library are designed to result in amplicons that are <800 bp, which is the optimal length for Illumina readout. This ensures that the sequence data is within the optimal range for Illumina's sequencing technology, allowing for better accuracy and read depth.</explanation> <answer>e</answer>.

\end{methodbox}


\section{Tables for Benchmark Models Evaluated and Accuracy Summary}

\begin{table}[htbp]
\centering
\renewcommand{\arraystretch}{1.2}
\setlength{\tabcolsep}{15pt}
\begin{tabular}{lll}
\toprule
\textbf{Model} & \textbf{Type} & \textbf{Version} \\
\midrule
\rowcolor{yellow!30}
\textbf{API Evaluated Commercial Models} & & \\
DeepSeek-R1 & Commercial & 01-20-25 \\
DeepSeek-V3 & Commercial & 03-25-25 \\
Claude-3.7-sonnet & Commercial & 02-19-25 \\
Claude-3.5-haiku & Commercial & 10-22-24 \\
Gemini-2.5-Pro & Commercial & exp-03-25 \\
Gemini-2.0-Flash & Commercial & exp-01-21 \\
GPT-4.1 & Commercial & 04-14-25 \\
o1 & Commercial & 12-17-24 \\
GPT-4o & Commercial & 11-20-24 \\
GPT-4o-mini & Commercial & 07-18-24 \\
\midrule
\rowcolor{yellow!30}
\textbf{Base Models} & & \\
Llama-3.1-8B & Open-Source & Instruct \\
Phi-4-mini & Open-Source & Instruct \\
Gemma2-9B & Open-Source & Instruct \\
Qwen2.5-7B & Open-Source & Instruct \\
Qwen2.5-1.5B & Open-Source & Instruct \\
\bottomrule
\end{tabular}
\vspace{0.4em}
\caption{Summary of models evaluated in this study, including their sources and versions.}
\label{tab:model-summary}
\end{table}

\begin{table}[h]
\centering
\rowcolors{2}{yellow!30}{white}
\begin{tabular}{lccc}
\toprule
\textbf{Model} & \textbf{Commercial (\%)} & \textbf{Before RL (\%)} & \textbf{After RL (\%)} \\
\midrule
\multicolumn{4}{c}{\textit{Commercial Models}} \\
\midrule
DeepSeek-R1 & \textbf{78.97} & -- & -- \\
DeepSeek-V3 & 72.92 & -- & -- \\
Claude-3.7-sonnet & 76.40 & -- & -- \\
Claude-3.5-haiku & 70.80 & -- & -- \\
Gemini-2.5-Pro & 77.46 & -- & -- \\
Gemini-2.0-Flash & 70.65 & -- & -- \\
gpt-4.1 & 73.83 & -- & -- \\
o1 & 71.41 & -- & -- \\
gpt-4o & 68.99 & -- & -- \\
gpt-4o-mini & 62.78 & -- & -- \\
\midrule
\multicolumn{4}{c}{\textit{Fine-tuned Models (Before vs. After RL)}} \\
\midrule
LLaMA-3.1-8B & -- & 55.98 & 71.10 \\
Phi-4-mini & -- & 62.18 & 74.28 \\
Gemma2-9b & -- & 57.64 & 72.92 \\
Qwen2.5-7B & -- & 58.85 & \textbf{76.85} \\
Qwen2.5-1.5B & -- & 48.41 & 65.66 \\
\midrule
\rowcolor{red!10}
\textbf{RL-Router Model} & -- & -- & \textbf{81.07} \\
\bottomrule
\end{tabular}
\vspace{0.4em}
\caption{Genome-Bench Accuracy Comparison of Commercial models, Base Models, RL-Tuned Models, and Router Model.}
\end{table}

\end{document}